\documentclass[10pt,twocolumn,letterpaper]{article}

\usepackage[pagenumbers]{cvpr} 
    
\usepackage[accsupp]{axessibility}

\usepackage{graphicx}
\usepackage{amsmath}
\usepackage{amssymb}
\usepackage{booktabs}

\usepackage{makecell}
\usepackage{colortbl}
\usepackage{url}
\usepackage{float}
\usepackage{subcaption}
\usepackage{graphicx}
\usepackage{wrapfig}
\usepackage{xcolor} 
\usepackage{framed}
\usepackage{color}
\usepackage{enumitem}
\usepackage{multirow}
\usepackage{arydshln} 
\usepackage{booktabs}
\usepackage{bbding}
\usepackage{pifont}

\makeatletter
\@namedef{ver@everyshi.sty}{}
\makeatother
\usepackage{tikz}

\definecolor{mediumpurple}{RGB}{30,144,255}
\definecolor{thistle}{RGB}{216, 191, 216}
\definecolor{lightgreen}{RGB}{188,238,104}
\definecolor{lightblue}{RGB}{191,239,255}
\definecolor{lightcoral}{RGB}{240,128,128}
\definecolor{pink}{RGB}{255,192,203}

\newcommand{\mixshape}{\raisebox{0.5pt}{\tikz\fill[thistle] (0,0) circle (.8ex);}}
\newcommand{\proshape}{\raisebox{0.5pt}{\tikz\fill[lightcoral] (0,0) circle (.8ex);}}
\newcommand{\fineshape}{\raisebox{0.5pt}{\tikz\fill[lightblue] (0,0) circle (.8ex);}}
\newcommand{\ensshape}{\raisebox{0.5pt}{\tikz\fill[pink](0,0) circle (.8ex);}}
\newcommand{\stshape}{\raisebox{0.5pt}{\tikz\fill[lightgreen](0,0) circle (.8ex);}}

\usepackage[pagebackref,breaklinks,colorlinks]{hyperref}

\usepackage[capitalize]{cleveref}
\crefname{section}{Sec.}{Secs.}
\Crefname{section}{Section}{Sections}
\Crefname{table}{Table}{Tables}
\crefname{table}{Tab.}{Tabs.}

\def\cvprPaperID{7218} 
\def\confName{CVPR}
\def\confYear{2023}

\begin{document}

\title{Frame Flexible Network}

\author{%
  Yitian Zhang$^{1}$\ \
  Yue Bai$^{1}$\ \ \
  Chang Liu$^{1}$\ \ \ 
  Huan Wang$^{1}$\ \ \ 
  Sheng Li$^{2}$\ \ \ 
  Yun Fu$^{1}$\\
    $^{1}$Northeastern University \quad $^{2}$University of Virginia\\
  {\tt\small \{zhang.yitian, bai.yue, liu.chang6, wang.huan\}@northeastern.edu}\\
  {\tt\small shengli@virginia.edu, yunfu@ece.neu.edu}
}

\maketitle

\begin{abstract}
Existing video recognition algorithms always conduct different training pipelines for inputs with different frame numbers, which requires repetitive training operations and multiplying storage costs.
If we evaluate the model using other frames which are not used in training, we observe the performance will drop significantly (see Fig.~\ref{fig:Misalignment}), which is summarized as Temporal Frequency Deviation phenomenon.
To fix this issue, we propose a general framework, named Frame Flexible Network (FFN), which not only enables the model to be evaluated at different frames to adjust its computation, but also reduces the memory costs of storing multiple models significantly.
Concretely, FFN integrates several sets of training sequences, involves Multi-Frequency Alignment (MFAL) to learn temporal frequency invariant representations, and leverages Multi-Frequency Adaptation (MFAD) to further strengthen the representation abilities.
Comprehensive empirical validations using various architectures and popular benchmarks solidly demonstrate the effectiveness and generalization of FFN (e.g., 7.08/5.15/2.17$\%$ performance gain at Frame 4/8/16 on Something-Something V1 dataset over Uniformer). 
Code is available at \href{https://github.com/BeSpontaneous/FFN}{https://github.com/BeSpontaneous/FFN.}
\end{abstract}

\section{Introduction}

The growing number of online videos boosts the research on video recognition, laying a solid foundation for deep learning which requires massive data.
Compared with image classification, video recognition methods need a series of frames to represent the video which scales the computation. Thus, the efficiency of video recognition methods has always been an essential factor in evaluating these approaches.
One existing direction to explore efficiency is designing lightweight networks~\cite{howard2017mobilenets, zhang2018shufflenet} which are hardware friendly.
Even if they increase the efficiency with an acceptable performance trade-off, these methods cannot make further customized adjustments to meet the dynamic-changing resource constraint in real scenarios.
In community, there are two lines of research being proposed to resolve this issue.
The first one is to design networks that can execute at various depths~\cite{huang2017multi} or widths~\cite{yu2018slim} to adjust the computation from the model perspective.
The other line of research considers modifying the resolutions of input data~\cite{li2020learning,yang2020mutualnet} to accommodate the cost from the data aspect.
However, these methods are carefully designed for 2D CNNs, which may hinder their applications on video recognition where 3D CNNs and Transformer methods are crucial components.

\begin{figure}[t]
    \centering
    \begin{subfigure}[b]{0.23 \textwidth}
           \centering
           \includegraphics[width=\textwidth]{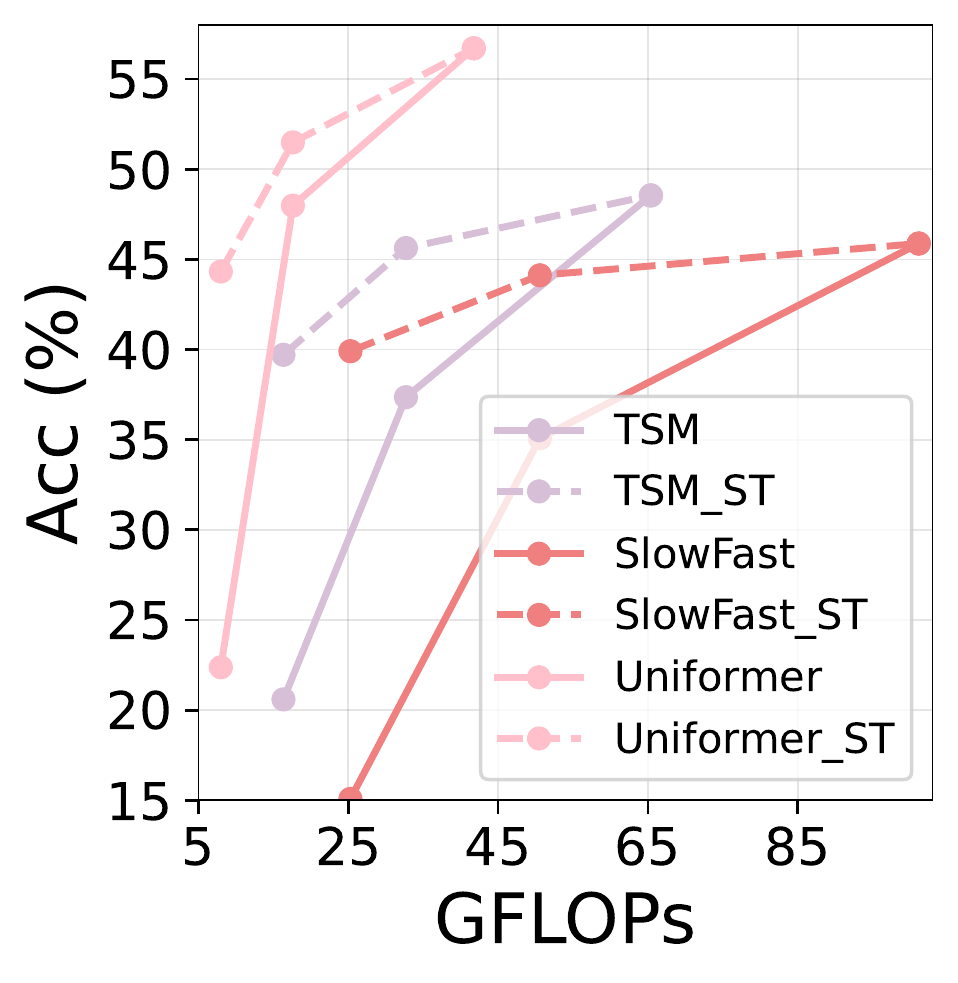}
           \vskip -0.05in
            \caption{Temporal Frequency Deviation phenomenon exists in various video recognition architectures.}
    \end{subfigure}
    \hfill
    \begin{subfigure}[b]{0.23 \textwidth}
            \centering
            \includegraphics[width=\textwidth]{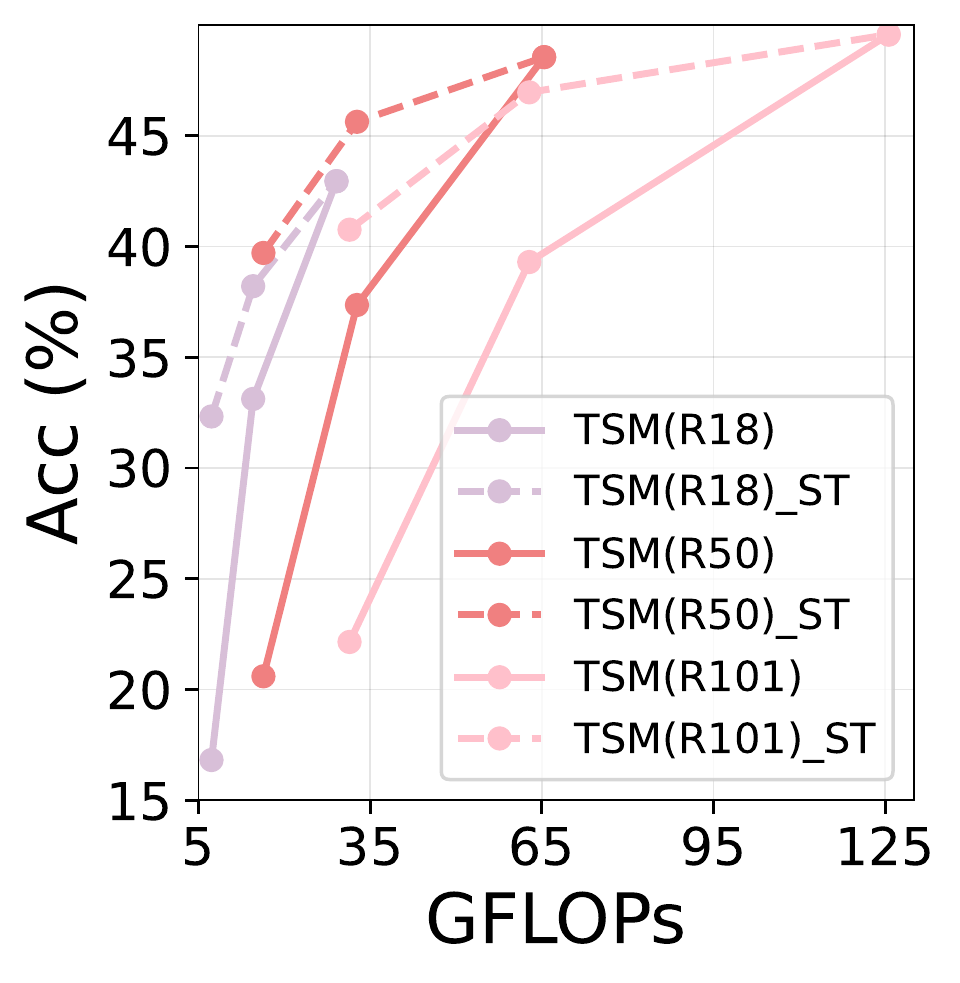}
           \vskip -0.05in
            \caption{Temporal Frequency Deviation phenomenon exists in different depths of deep networks.}
    \end{subfigure}
    \vskip -0.1in
    \caption{Temporal Frequency Deviation phenomenon widely exists in video recognition. All methods are trained with high frame number and evaluated at other frames to compare with Separated Training (ST) which individually trains the model at different frames on Something-Something V1 dataset.}
\label{fig:Misalignment}
\vskip -0.2in
\end{figure}

Different from image-related tasks, we need to sample multiple frames to represent the video, and the computational costs will grow proportionally to the number of sampled frames.
Concretely, standard protocol trains the same network with different frames separately to obtain multiple models with different performances and computations. This brings challenges to applying these networks on edge devices as the parameters will be multiplied if we store all models, and downloading and offloading models to switch them will cost non-negligible time. Moreover, the same video may be sampled at various temporal rates on different platforms, 
employing a single network that is trained at a certain frame number for inference cannot 
resist the variance of frame numbers in real scenarios.

Training the model with a high frame number (i.e., high temporal frequency) and directly evaluating it at fewer frames (i.e., low temporal frequency) to adjust the cost is a naive and straightforward solution.
To test its effectiveness, we compare it with Separated Training (ST) which trains the model at different temporal frequency individually and tests it with the corresponding frame.
We conduct experiments on 2D-network TSM~\cite{lin2019tsm}, 3D-network SlowFast~\cite{feichtenhofer2019slowfast} and Transformer-network Uniformer~\cite{li2022uniformer}, and find obvious performance gaps between the inference results and ST 
from Fig.~\ref{fig:Misalignment}, 
which means these methods will exhibit significantly inferior performance if they are not evaluated at the frame number used in training. 
Further, we conduct the same experiments on different depths of deep networks and a similar phenomenon appears.
We denote this generally existing phenomenon as Temporal Frequency Deviation.

The potential reason for Temporal Frequency Deviation has been explored in Sec.~\ref{sec:shift} and briefly summarized as the shift in normalization statistics. To address this issue, we propose a general framework, named Frame Flexible Network (FFN), which only requires one-time training, but can be evaluated at multiple frame numbers with great flexibility. We import several input sequences with different sampled frames to FFN during training and 
propose Multi-Frequency Alignment (MFAL) to learn the temporal frequency invariant representations for robustness towards frame change. 
Moreover, we present Multi-Frequency Adaptation (MFAD) to further strengthen the representation abilities of the sub-networks which helps FFN to exhibit strong performance at different frames during inference.

Although normalization shifting problem~\cite{yu2018slim,yu2019universally} and resolution-adaptive networks~\cite{li2020learning,yang2020mutualnet} have been studied, we stress that designing frame flexible video recognition frameworks 
to accommodate the costs and save parameters
is non-trivial and has practical significance for the following reasons. First, prior works~\cite{li2020learning,yang2020mutualnet} carefully analyzed the detailed structure of 2D convolutions in order to privatize the weights for different scale images. While our method does not touch the specific design of 
the spatial-temporal modeling components
and shares their weights for inputs with different frames. This procedure not only enables our method to be easily applied to various architectures (2D/3D/Transformer models), but also enforces FFN to learn temporal frequency invariant representations. 
Second, it is, indeed, a common practice to conduct Separated Training (ST) in video recognition, which needs multiplying memory costs to store individual models, and the models are hard to resist the variance in temporal frequency which limits their applications in actual practice. While FFN provides a feasible solution to these challenges which significantly reduces the memory costs of storing multiple models and can be evaluated at different frames to adjust the cost with even higher accuracy compared to ST.

With the proposed framework, we can resolve Temporal Frequency Deviation 
and enable these methods to adjust their computation based on the current resource budget by sampling different frames, trimming the storage costs of ST remarkably. Moreover, we provide a naive solution that enables FFN to be evaluated at any frame and increases its flexibility during inference.  Validation results prove that FFN outperforms ST even at frames that are not used in training. The contributions are summarized as follows:
\begin{itemize}[itemsep=0pt,topsep=0pt,parsep=2pt]
    \item We reveal the phenomenon of \emph{Temporal Frequency Deviation} that widely exists in video recognition. It is detailedly analyzed and practically inspires our study.
    \item We propose a general framework \emph{Frame Flexible Network} (FFN) to resolve Temporal Frequency Deviation. 
    We design Multi-Frequency Alignment (MFAL) to learn temporal frequency invariant representations and present Multi-Frequency Adaptation (MFAD) to further strengthen the representation abilities.
    \item Comprehensive empirical validations show that FFN, which only requires one-shot training, can adjust its computation by sampling different frames and outperform Separated Training (ST) at different frames on various architectures and datasets, reducing the memory costs of storing multiple models significantly.
\end{itemize}

\section{Related Work}

\noindent\textbf{Video Recognition} has been extensively explored in recent years and we can summarize the methods into three categories based on their architectures: 1) 2D networks: these methods~\cite{wang2016temporal,lin2019tsm,li2020tea,wang2021tdn} utilize 2D CNNs as the backbone and specifically design temporal modeling module for spatial-temporal modeling. 2) 3D networks: a straightforward solution for video recognition is to utilize 3D convolutions~\cite{tran2015learning,carreira2017quo,feichtenhofer2019slowfast} which naturally consider the temporal information in frame sequences. 3) Transformer networks: based on Vision Transformers~\cite{dosovitskiy2020image,liu2021swin}, many approaches~\cite{fan2021multiscale,liu2022video,li2022uniformer} have been proposed recently for spatial-temporal learning and shown powerful performance.

\noindent\textbf{Training-testing Discrepancy} widely exists in many scenarios of deep learning. FixRes~\cite{touvron2019fixing} discovers the deviation of image resolutions between training and testing. Based on this observation, there are methods~\cite{li2020learning,yang2020mutualnet} being designed to train a  universal network to fit the images at different resolutions and ~\cite{yang2021mutualnet} further extended 
this idea to 3D CNNs. Slimmable Neural Networks~\cite{yu2018slim,yu2019universally} train a shared network which can adjust its width to meet the resource constraints during inference. Different from these prior works, our work is motivated by Temporal Frequency Deviation in video recognition. This finding is essential as frame sampling is a necessary step for all methods and former procedures train the network with different frames individually which is parameter-inefficient and memory-consuming.

\noindent\textbf{Parameter-efficient Transfer Learning} has aroused researchers' attention in NLP because of the arising of large-scale pre-trained language models. An important research line is to design task-specific adapters~\cite{pfeiffer2020adapterfusion,pfeiffer2020adapterhub} to achieve parameter-efficient. Recently, the idea of adapters has been extended to vision tasks as well and shown favorable performance~\cite{sung2022vl,pan2022st,zhang2022minivit}. In this work, instead of focusing on tuning from large-scale pre-trained models, we present Multi-Frequency Adaptation (MFAD) to increase the representation abilities of sub-networks.

\noindent\textbf{Dynamic Networks} have been widely studied for efficient video recognition in recent years. Some methods~\cite{zhang2022look,wu2020dynamic,korbar2019scsampler} dynamically sample salient frames to reduce temporal redundancy for less cost, while others mainly focus on reducing spatial redundancy by adaptively processing frames with different resolutions~\cite{meng2020ar} or cropping the most salient regions~\cite{wang2021adaptive} for each frame. Note that these methods are designed to adaptively process every video (e.g., skip frames, crop patches) for efficiency and also requires repetitive training to obtain models with different computation. Our work aims to train a model which can be evaluated at different frames to adjust the costs and reduce the parameters of storing multiple models, while the mentioned dynamic networks do not solve this problem.

\begin{figure}[t]
    \centering
    \begin{subfigure}[b]{0.23 \textwidth}
            \centering
            \includegraphics[width=\textwidth]{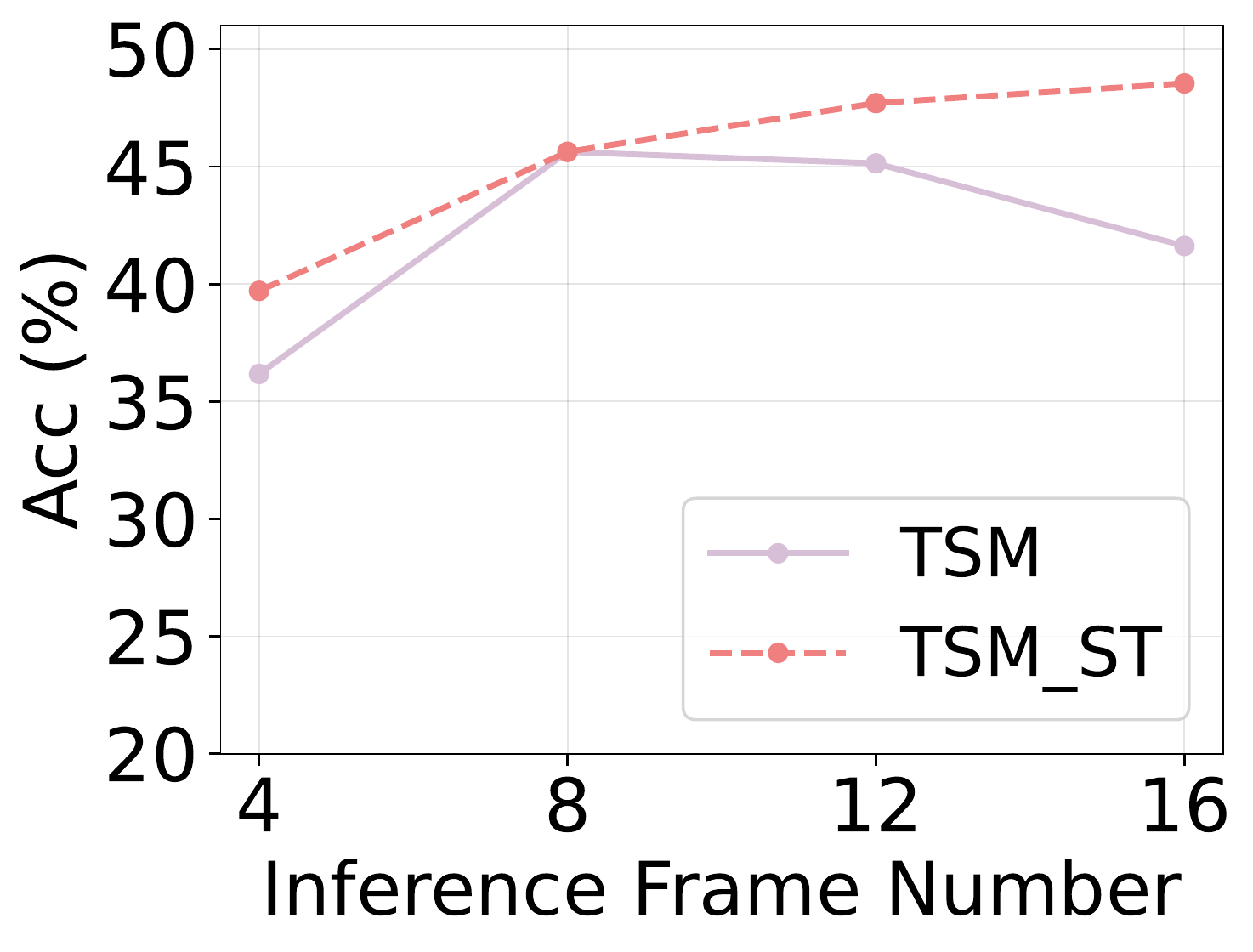}
           \vskip -0.05in
            \caption{8 Frame Training.}
    \end{subfigure}
    \begin{subfigure}[b]{0.23 \textwidth}
            \centering
            \includegraphics[width=\textwidth]{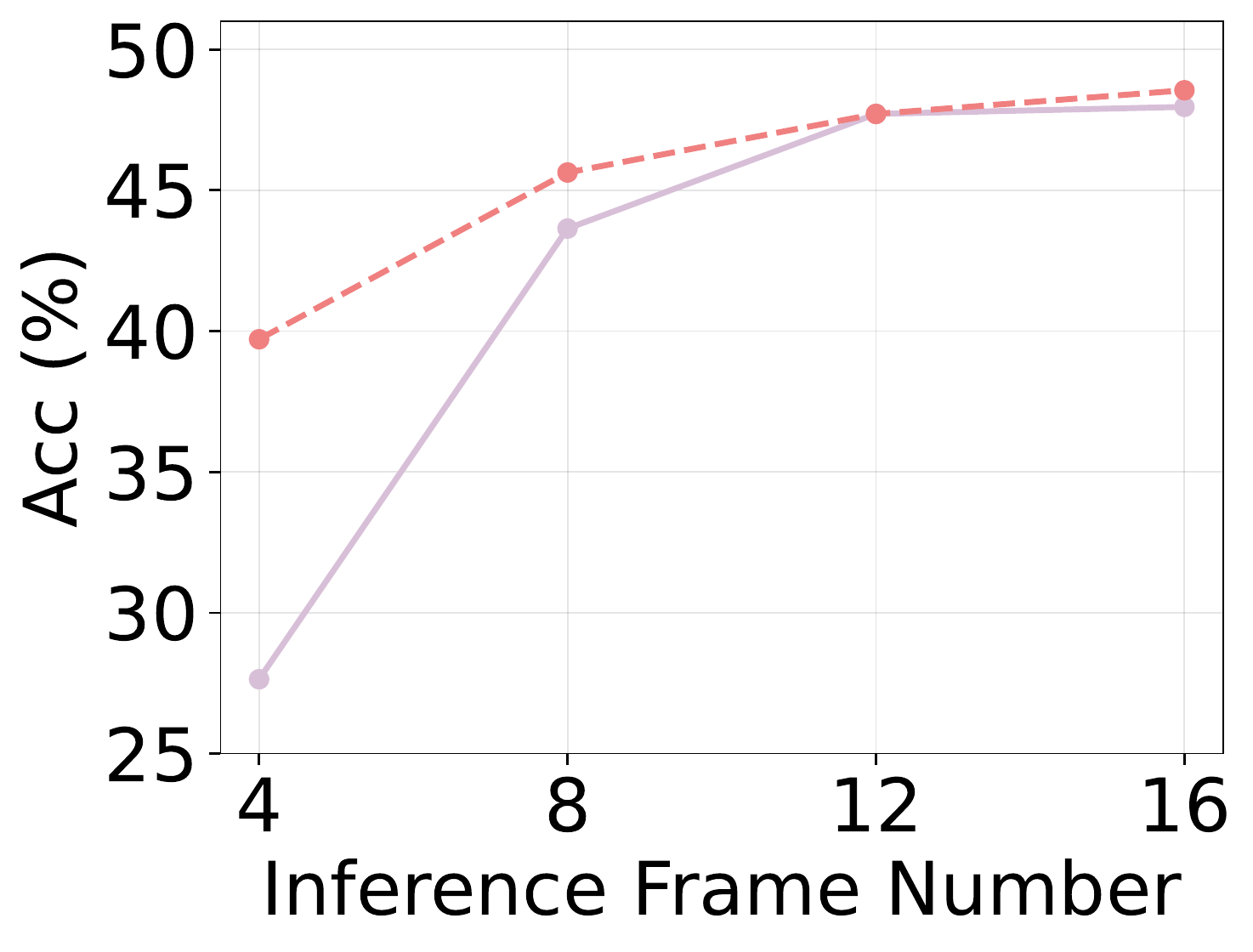}
           \vskip -0.05in
            \caption{12 Frame Training.}
    \end{subfigure}
    \vskip -0.1in
    \caption{Nearby Alleviation phenomenon. TSM model is trained at 8 Frame and 12 Frame separately on Something-Something V1 dataset and will be evaluated at other frames.}
\label{fig:local}
\vskip -0.2in
\end{figure}

\section{Temporal Frequency Deviation}

\noindent\textbf{Nearby Alleviation.}\label{sec:near} We can observe Temporal Frequency Deviation phenomenon when the models are trained with high frame numbers but evaluated at fewer frames from Fig.~\ref{fig:Misalignment}. To step further, we train TSM~\cite{lin2019tsm} at 8/12 Frame and evaluate them at other frames. It is shown in Fig.~\ref{fig:local} that there are performance gaps for both models if it is not evaluated with the same frame number which is used in training. Particularly, the discrepancies vary in terms of the value and the performance gap is smaller if the inference frame is close to the training frame number. We denote this phenomenon as Nearby Alleviation because Temporal Frequency Deviation is less severe at nearby frames.

\begin{figure}[t]
    \centering
    \begin{subfigure}[b]{0.23 \textwidth}
           \centering
           \includegraphics[width=\textwidth]{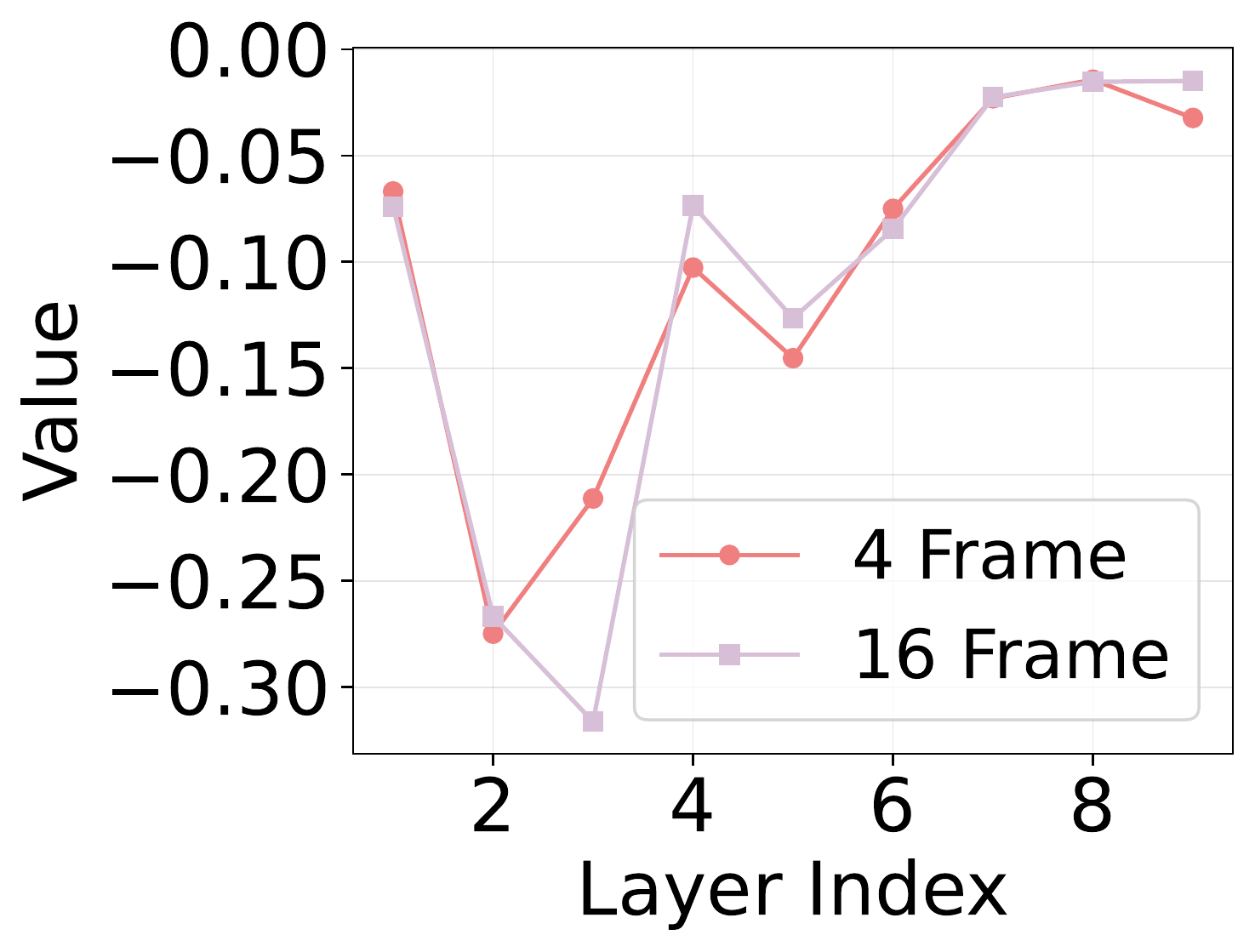}
           \vskip -0.05in
            \caption{Mean: $\mu$}
    \end{subfigure}
    \hfill
    \begin{subfigure}[b]{0.23 \textwidth}
            \centering
            \includegraphics[width=\textwidth]{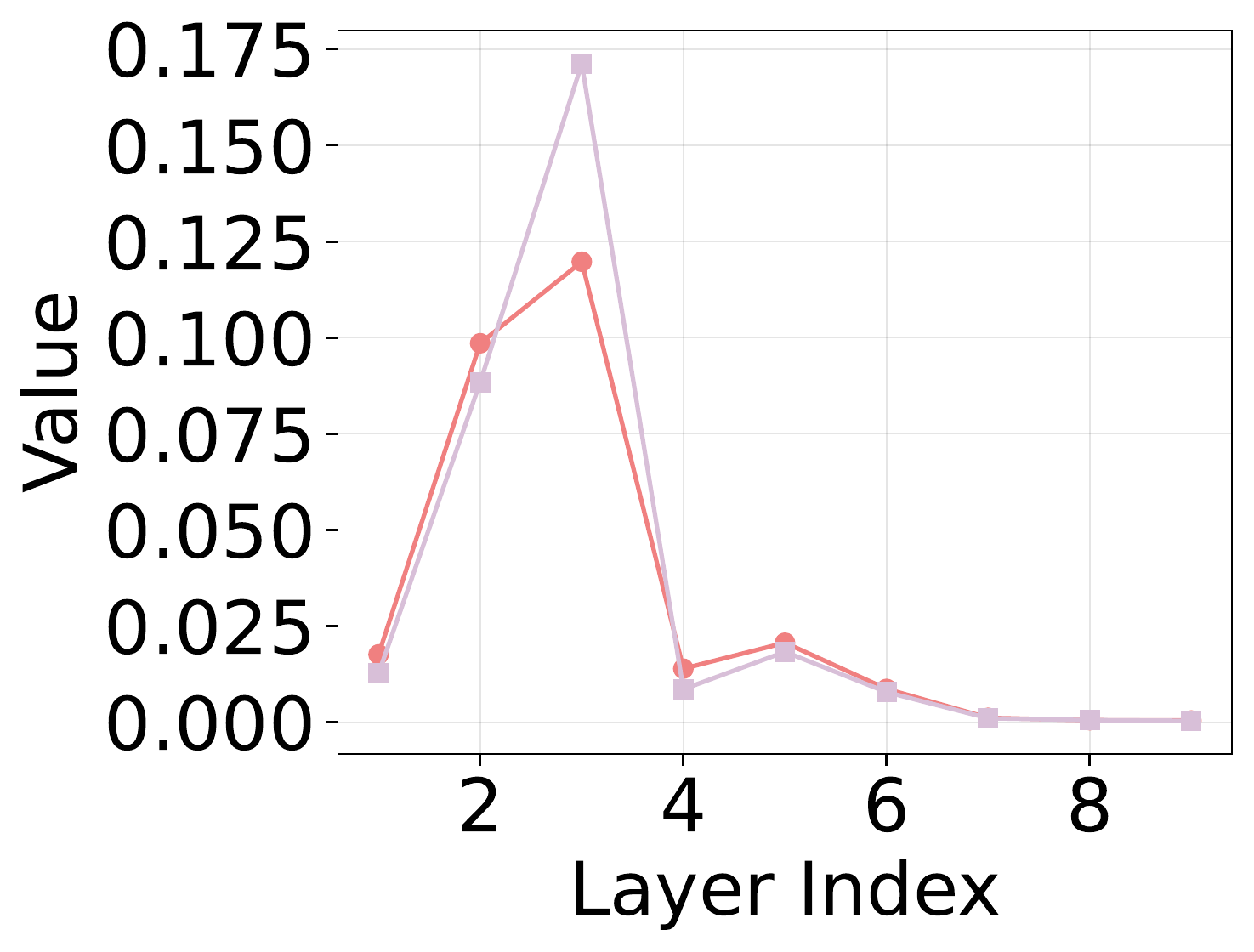}
           \vskip -0.05in
            \caption{Variance: $\sigma^{2}$}
    \end{subfigure}
    \vskip -0.1in
    \caption{Batch Normalization statistics at various layers. TSM models are trained at 4 Frame and 16 Frame separately, and the statistics are calculated from the fourth stage of ResNet-50.}
\label{fig:bn}
\vskip -0.2in
\end{figure}

\begin{figure*}[t]
\begin{center}
\scalebox{1.0}{\includegraphics[width=\textwidth]{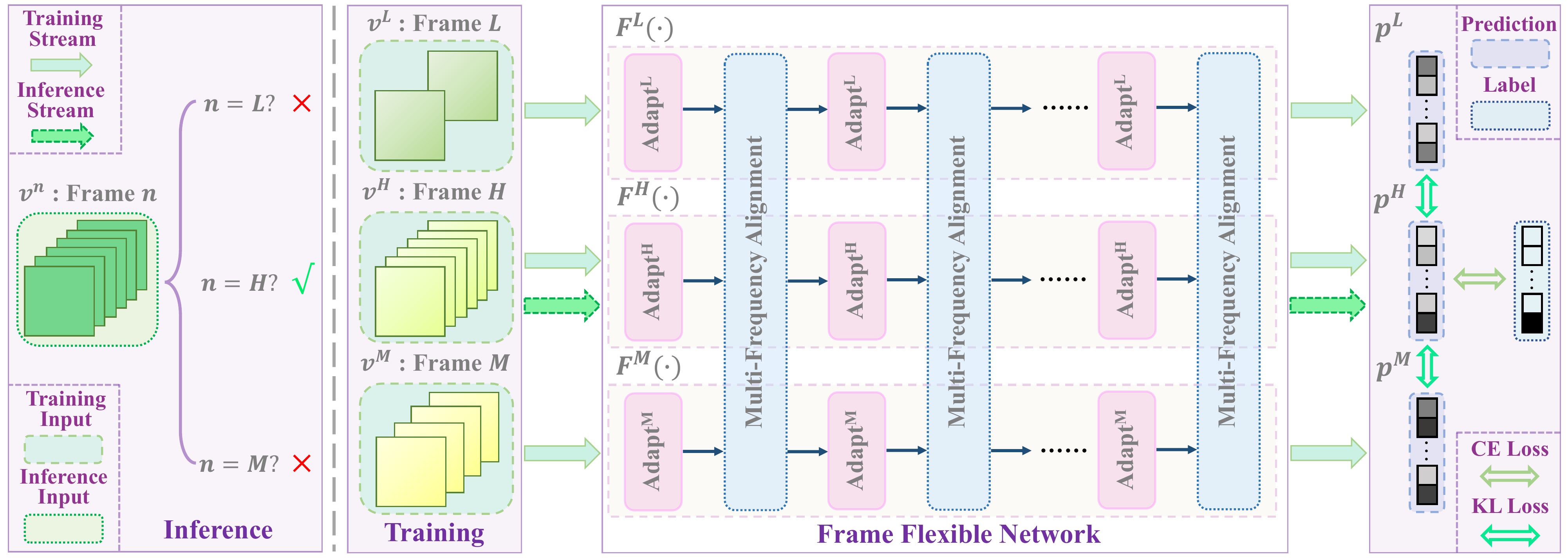}}
\end{center}
\vskip -0.2in
\caption{
Illustration of Frame Flexible Network (FFN). During training, given inputs with different temporal frequency $v^{L}$, $v^{M}$ and $v^{H}$, we propose Multi-Frequency Alignment which involves Weight Sharing and Temporal Distillation for temporal frequency invariant learning. Besides, we present Multi-Frequency Adaptation to fit the temporal invariant features to different sub-networks to further increase the representation abilities. During inference, we activate the sub-network which has the corresponding frame number with the input.
}
\label{fig:method}
\vskip -0.2in
\end{figure*}

\noindent\textbf{Normalization Shifting.}\label{sec:shift} 
Prior works~\cite{yu2018slim,yu2019universally} have studied the problem of normalization shifting in image classification.
Specifically, 
when switching the widths of networks, different numbers of channels will lead to different means and variances of the aggregated features, leading to inconsistency in feature aggregation.

While we do not consider the adjustment in model structure, the problem is whether the difference in frame numbers will cause normalization shifting.
If we train the model with $v^{H}$ which has high temporal frequency and evaluate it with low temporal frequency $v^{L}$, the input of Batch Normalization (BN) will be the intermediate feature $x^{L}$ and the corresponding output is:
\begin{equation}
    {y^{L}}' = \gamma^{H} \frac{x^{L} - \mu^{H}}{\sqrt{{\sigma^{H}}^{2} + \epsilon }} + \beta^{H}, 
\end{equation}
where $\mu^{H}$, ${\sigma^{H}}^{2}$ are calculated from the data distribution of $v^{H}$, and $\gamma^{H}$, $\beta^{H}$ are learnt at the training process with $v^{H}$. 
We calculate the statistics of the models trained with $v^{L}$ and $v^{H}$ separately and show it in Fig.~\ref{fig:bn}. We can observe a discrepancy of BN statistics at different frame numbers. Note that $\mu$ and $\sigma^{2}$ are data-dependent which means that the divergence lies in data intrinsically. Thus, we conjecture that the discrepancy of BN statistics at different frames is an essential factor which leads to Temporal Frequency Deviation. Layer Normalization (LN)~\cite{ba2016layer} has been widely used in Transformer-based models and its statistics are calculated in a similar way with BN which is related to the data distribution. Therefore, we believe the discrepancy of LN statistics is also one of the reasons for Temporal Frequency Deviation on Transformer-based models.

\section{Frame Flexible Network}

In this section, we first present the training and inference paradigms of Frame Flexible Network (FFN).
Then, we propose Multi-Frequency Alignment which is composed of Weight Sharing and Temporal Distillation to learn temporal frequency invariant representations. Further, we introduce Multi-Frequency Adaptation which fits the frequency invariant features to different sub-networks and further increases their representation abilities.
Note that FFN is a general framework which can be built on different architectures (shown in Sec.~\ref{sec:structures}) and we just take CNN based method as an example in this part for easier description.

\subsection{Framework}

The goal of our work is to present a method which can be evaluated at multiple frames and exhibits similar or even better performance compared to Separated Training (ST).
Based on the analysis in Sec.~\ref{sec:near}, Temporal Frequency Deviation will be less severe if the model is evaluated at nearby frames which are used in training. Therefore, we decide to import several sequences with different sampled frames to FFN shown in Fig.~\ref{fig:method}. Consider video $v$ which is sampled at increasing frame numbers $L$, $M$ and $H$, we can obtain $v^{L}$, $v^{M}$ and $v^{H}$ with temporal frequency of Low, Medium and High, respectively. These three sequences will be utilized at training phase to construct three sub-networks $F^{L}\left (\cdot  \right )$, $F^{M}\left (\cdot  \right )$ and $F^{H}\left (\cdot  \right )$ accordingly. As for the inference paradigm, we will activate the sub-network which has the corresponding frame number with the input. 
In this manner, we build the computational stream that enables FFN to be evaluated with different frames during inference and adjust the computational costs accordingly.

\begin{figure*}[t]
\begin{center}
\scalebox{1}{\includegraphics[width=\textwidth]{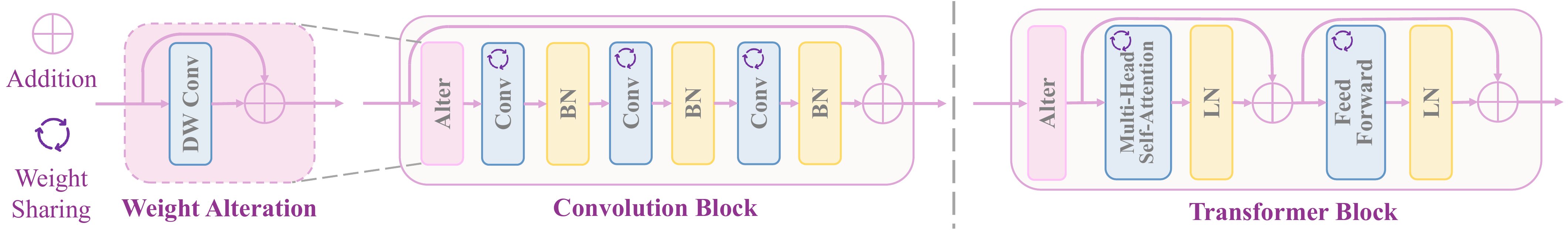}}
\end{center}
\vskip -0.2in
\caption{Specific designs of Weight Alteration, Convolution Block, and Transformer Block in Frame Flexible Network (FFN). Weight Alteration is a Depth-wise convolution layer with a residual structure and we insert it into each Convolution Block and Transformer Block.}
\label{fig:block}
\vskip -0.2in
\end{figure*}

\subsection{Multi-Frequency Alignment}

Prior resolution-adaptive networks~\cite{li2020learning,yang2020mutualnet} carefully privatize the weights for 2D convolutions to 
learn the scale-aware representations for inputs with different resolutions.
Recently, there are several works~\cite{yang2020video,singh2021semi} being proposed to maximize the mutual information of the same video at different temporal frequency for contrastive learning in video recognition. The core idea is that the same video instance with different speeds should share high similarity in terms of their discriminative semantics.
Inspired by these works, we propose Multi-Frequency Alignment (MFAL) which leverages Weight Sharing and Temporal Distillation to efficiently expand the network and enforce the model to learn temporal frequency invariant representations.

\noindent\textbf{Weight Sharing.} Given video $v$, we have $v^{L}$, $v^{M}$, and $v^{H}$ with an increased temporal frequency and decreased action speed because of the difference in sampled frames. 
We share the weights of convolutions and classifier across the three sub-networks in order to find a group of parameters $\theta$ that mutually model the spatial-temporal relationships for inputs with different temporal frequency:
\begin{equation}
    p^{*} = F^{*}\left (v^{*};\theta  \right ),
\end{equation}
where $* \in \{L, M, H\}$ and $p$ stands for the predictions. 
Compared to specialized convolutions, Weight Sharing is parameter-efficient as it only stores one set of weights which can be applied to different input frames. Moreover, it exhibits great potential for better performance (shown in Tab.~\ref{tab:ablation}) as it will enforce the model to learn temporal frequency invariant representations which implicitly provides the prior knowledge that the same video with different temporal frequency belongs to the same class, making the model robust to temporal frequency variance.

\noindent\textbf{Temporal Distillation.} In most cases, video recognition models trained with $v^{H}$ have better performance as the network will have access to more information of the original video. Therefore, we consider $p^{H}$ to be the most `\textit{accurate}' prediction among the three as $v^{H}$ has the most sampled frames. Applying Cross-Entropy loss on $p^{H}$, we can update the parameters of $F^{H}\left (\cdot  \right )$ by:
\begin{equation}
	\mathcal{L}_{CE} = -\sum_{k=1}^{K} \hat{y}_{k}\log \left ( p_{k}^{H} \right ),\label{equa:ce}
\end{equation}
where $\hat{y}_{k}$ is the one-hot label of class $k$ and there are $K$ classes in total. 
Directly calculating CE loss on $p^{L}$ and $p^{M}$ is a straightforward solution to update the parameters in $F^{L}\left (\cdot  \right )$ and $F^{M}\left (\cdot  \right )$, but it will lead to some problems. 
Firstly, the weights of convolutions are shared across three sub-networks and the optimal parameters for $v^{L}$ after optimization may not fit well to $v^{M}$ and $v^{H}$. Moreover, optimizing CE loss of $p^{L}$ and $p^{M}$ will lead to less favorable parameters of convolutions compared to only calculating Eq.~\ref{equa:ce} as their inputs contain less information compared to $v^{H}$ which may lead to inferior performance.

Consequently, we utilize KL divergence~\cite{kullback1997information} loss
to involve $p^{L}$ and $p^{M}$ in the computational graph and update the parameters of $F^{L}\left (\cdot  \right )$ and $F^{M}\left (\cdot  \right )$ using:
\begin{equation}
	\mathcal{L}_{KL} = -\sum_{k=1}^{K} p_{k}^{H}\log \left ( \frac{p_{k}^{M}}{p_{k}^{H}}\right ) -\sum_{k=1}^{K} p_{k}^{H}\log \left ( \frac{p_{k}^{L}}{p_{k}^{H}}\right ).\label{eq:distill}
\end{equation} 
As the weights of convolutions are shared across the three sub-networks, optimizing Eq.~\ref{eq:distill} will enforce the predictions of student ($p^{L}$ and $p^{M}$) and teacher ($p^{H}$) networks to be as similar as possible and transfer the good knowledge from $F^{H}\left (\cdot  \right )$ to $F^{L}\left (\cdot  \right )$ and $F^{M}\left (\cdot  \right )$. Considering the two losses in a uniform manner, we update the parameters of FFN by:
\begin{equation}
	\mathcal{L} = \mathcal{L}_{CE} + \lambda \cdot \mathcal{L}_{KL},
\end{equation}
where $\lambda$ is an introduced hyperparameter to balance the two terms and we simply let $\lambda=1$ in our implementations without fine-tuning the hyperparameter.

Considering Weight Sharing and Temporal Distillation uniformly, $\mathcal{L}_{CE}$ will provide inter-class supervisory information to enlarge the distance between videos belonging to different classes, and $\mathcal{L}_{KL}$ will further add intra-instance knowledge to the network training, i.e., $p^{L}$, $p^{M}$ and $p^{H}$ should share high similarity with each other as temporal frequency variance will not change the class of the video. 
In this way, we not only enforce FFN to learn temporal frequency invariant representations, but also promise it to be easily applied to different structures as we do not touch the specific design of inner spatial-temporal modeling modules.

\begin{table*}[t]
\centering
\scalebox{0.85}{\begin{tabular}{lccccccccc}
\toprule
\multirow{2}*{Method} & \multirow{2}*{Specification} & \multirow{2}*{Parameters} & \multicolumn{2}{c}{4 Frame (4F)} & \multicolumn{2}{c}{8 Frame (8F)} & \multicolumn{2}{c}{16 Frame (16F)} \\
\cmidrule(lr){4-5}\cmidrule(lr){6-7}\cmidrule(lr){8-9}
& & & Top-1 Acc. ($\%$) & GFLOPs & Top-1 Acc. ($\%$) & GFLOPs & Top-1 Acc. ($\%$) & GFLOPs \\
\midrule
TSM~\cite{lin2019tsm}  & - & 25.6M & 20.60 & 16.4 & 37.36 & 32.7 & 48.55 & 65.4 \\
\midrule
\mixshape{} TSM-Mixed  & $\rho=0.50$ & 25.6M & 27.89 & 16.4 & 41.07 & 32.7 & 48.44 & 65.4 \\
\mixshape{} TSM-Mixed  & $\rho=0.75$ & 25.6M & 30.43 & 16.4 & 42.56 & 32.7 & 47.81 & 65.4 \\
\proshape{} TSM-Proportional  & $\varrho=0.50$ & 25.6M & 37.56 & 16.4 & 44.82 & 32.7 & 45.37 & 65.4 \\
\proshape{} TSM-Proportional  & $\varrho=0.75$ & 25.6M & 32.06 & 16.4 & 43.15 & 32.7 & 47.14 & 65.4 \\
\fineshape{} TSM-Fine-tuning  & 16F$\rightarrow$4F & 25.6M & \textcolor{mediumpurple}{39.95} & 16.4 & 40.37 & 32.7 & 28.96 & 65.4 \\
\ensshape{} TSM-Ensemble  & - & 25.6$\times$3M & 35.88 & 16.4$\times$3 & \textcolor{mediumpurple}{46.25} & 32.7$\times$3 & 46.82 & 65.4$\times$3 \\
\stshape{} TSM-ST  & - & 25.6$\times$3M & 39.71 & 16.4 & 45.63 & 32.7 & \textcolor{mediumpurple}{48.55} & 65.4 \\
\midrule
TSM-FFN  & - & 25.7M & \textbf{42.85} \textcolor{mediumpurple}{(2.90$\uparrow$)} & 16.4 & \textbf{48.20} \textcolor{mediumpurple}{(1.95$\uparrow$)} & 32.8 & \textbf{50.79} \textcolor{mediumpurple}{(2.24$\uparrow$)} & 65.5 \\
\bottomrule
\end{tabular}}
\vskip -0.05in
\caption{Comparison with baseline methods on Something-Something V1 dataset. GFLOPs stands for the average computational cost to process a single video. The best results are bold-faced, the second best results are marked in color and the improvements are shown.}
\label{tab:ensemble}
\vskip -0.2in
\end{table*}

\subsection{Multi-Frequency Adaptation}

In the previous section, we propose MFAL to enforce FFN to learn temporal frequency invariant representations. Here, we present Multi-Frequency Adaptation (MFAD) to better fit the frequency invariant features to different sub-networks which 
further strengthen their representations.

According to our analysis in Sec.~\ref{sec:shift}, normalization shifting is one of the reasons which leads to Temporal Frequency Deviation. 
Formally, we denote the intermediate features for $v^{L}$, $v^{M}$ and $v^{H}$ as $x^{L}$, $x^{M}$ and $x^{H}$, respectively. Similar with~\cite{yu2018slim,yu2019universally}, we provide specialized normalization for different input sequences $v^{L}$, $v^{M}$ and $v^{H}$:
\begin{equation}
    y^{*} = \gamma^{*} \frac{x^{*} - \mu^{*}}{\sqrt{{\sigma^{*}}^{2} + \epsilon }} + \beta^{*},
\end{equation}
where $* \in \{L, M, H\}$, and private normalization will learn its own $\gamma$ and $\beta$ and calculate the corresponding $\mu$, $\sigma^{2}$ during training.
Note that this procedure introduces negligible computation and parameters as normalization operation is a simple transformation and its parameters are often less than 1$\%$ of the model size.

\noindent\textbf{Weight Alteration.} Though Weight Sharing is necessary for MFAL, 
it may be difficult to find a set of parameters to display strong representation ability at all frames without further adaptation. 
Considering a shared convolution with weights $W$, the outputs of different sequences are:
\begin{equation}
    y^{*} = W \otimes  x^{*},
\end{equation}
where $\otimes$ stands for convolution which applies the same transformation for 
inputs with different temporal frequency.
We propose to alter the shared weights of each sub-network to diversify the parameters and strengthen their representation abilities through the transformation:
\begin{equation}
    y^{*} = \phi^{*} \otimes W \otimes x^{*},\label{eq:wa}
\end{equation}
which can also be written as:
\begin{equation}
    y^{*} = W^{*} \otimes x^{*},\ \
    W^{*} = \phi^{*} \otimes W,
\end{equation}
where $\phi$ is a Depth-Wise convolution layer~\cite{chollet2017xception} at each Convolution Block which can covert the shared weights $W$ into diversified weights $W^{*}$.
In this way, we can increase the representation ability of FFN through a simple and efficient transformation.
Given that video recognition methods often use pre-trained models, we include the residual structure~\cite{he2016deep} to avoid the added module breaking the original computational graph of pre-trained models and restore their behaviors. Similarly, we also include Weight Alteration in Transformer Block and we choose the inserted location following~\cite{pan2022st} shown in Fig.~\ref{fig:block}. 
Note that Depth-Wise convolution is lightweight and adding it will introduce negligible parameters and computation.

\section{Experiments}

In this part, we validate Frame Flexible Network (FFN) on various architectures and benchmarks. First, we provide several baseline solutions and compare them with FFN. Further, we apply our method to different methods and datasets to prove its generalization ability.  Moreover, we provide a naive inference paradigm to enable FFN to be evaluated at any frame. 
Finally, we conduct detailed ablations and analyses to validate the effectiveness of our designs.

\subsection{Experiment Settings}

\begin{figure*}[t]
    \centering
    \begin{subfigure}[b]{0.24 \textwidth}
          \centering
          \includegraphics[width=\textwidth]{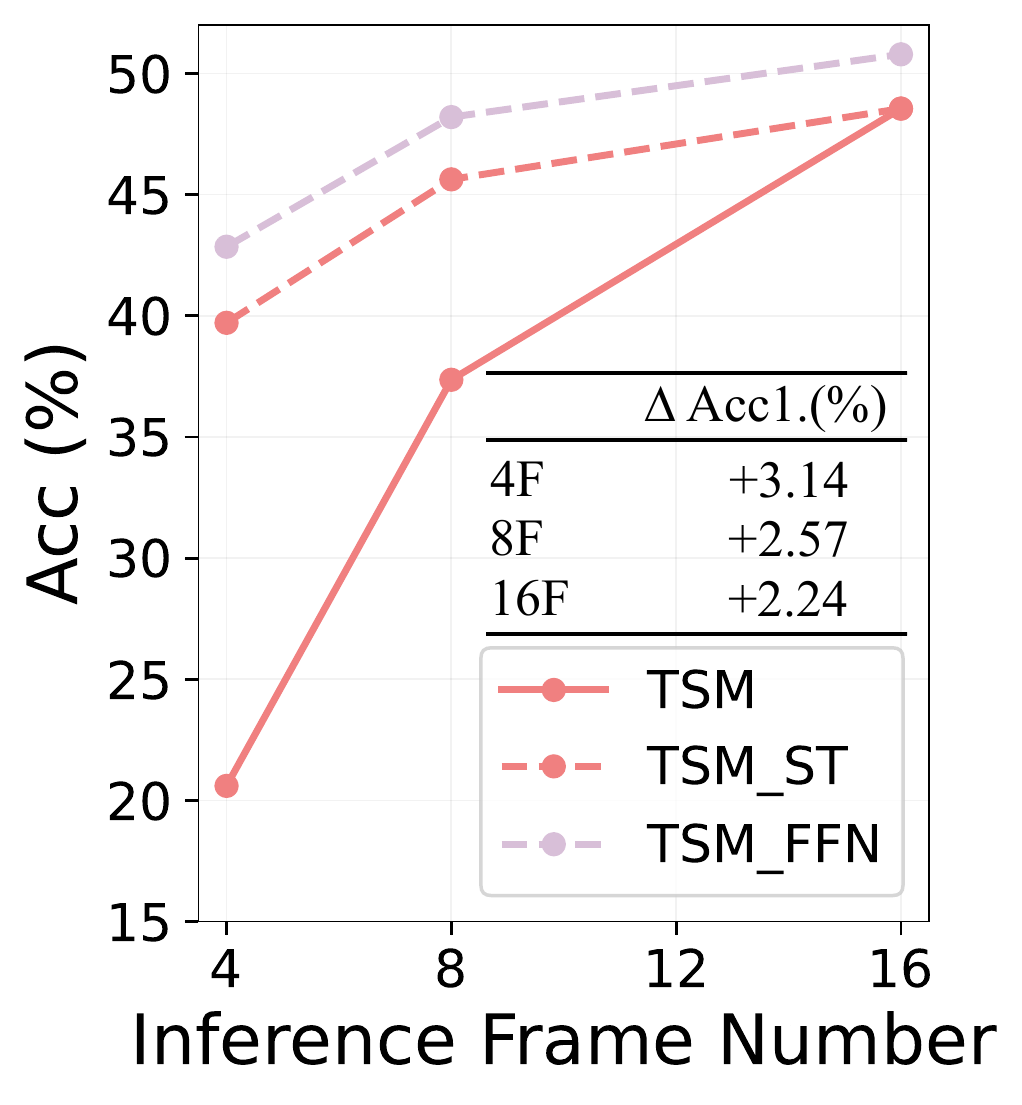}
          \vskip -0.05in
            \caption{Results on TSM.}
    \end{subfigure}
    \hfill
    \begin{subfigure}[b]{0.24 \textwidth}
            \centering
            \includegraphics[width=\textwidth]{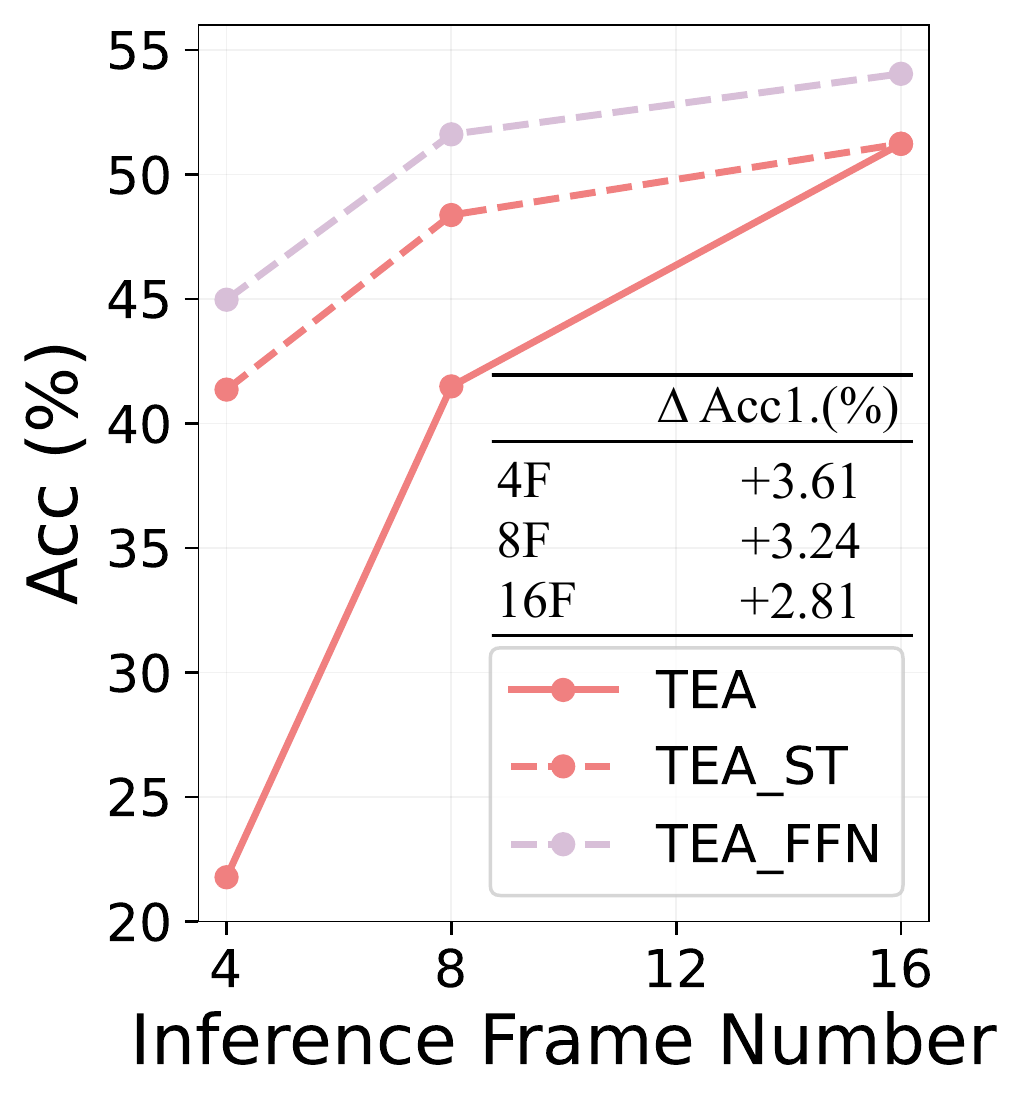}
          \vskip -0.05in
            \caption{Results on TEA.}
    \end{subfigure}
    \hfill
    \begin{subfigure}[b]{0.24 \textwidth}
          \centering
          \includegraphics[width=\textwidth]{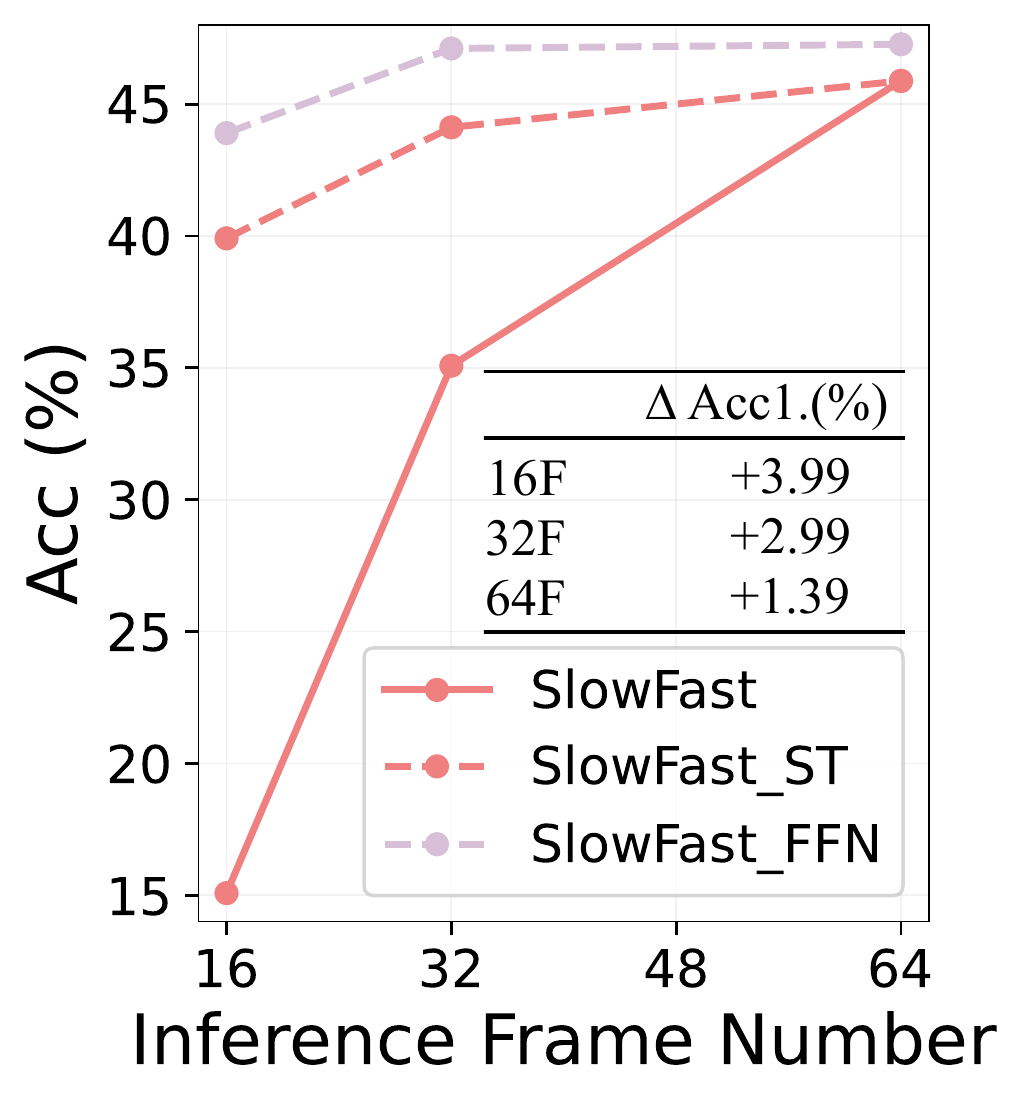}
          \vskip -0.05in
            \caption{Results on SlowFast.}
    \end{subfigure}
    \hfill
    \begin{subfigure}[b]{0.24 \textwidth}
            \centering
            \includegraphics[width=\textwidth]{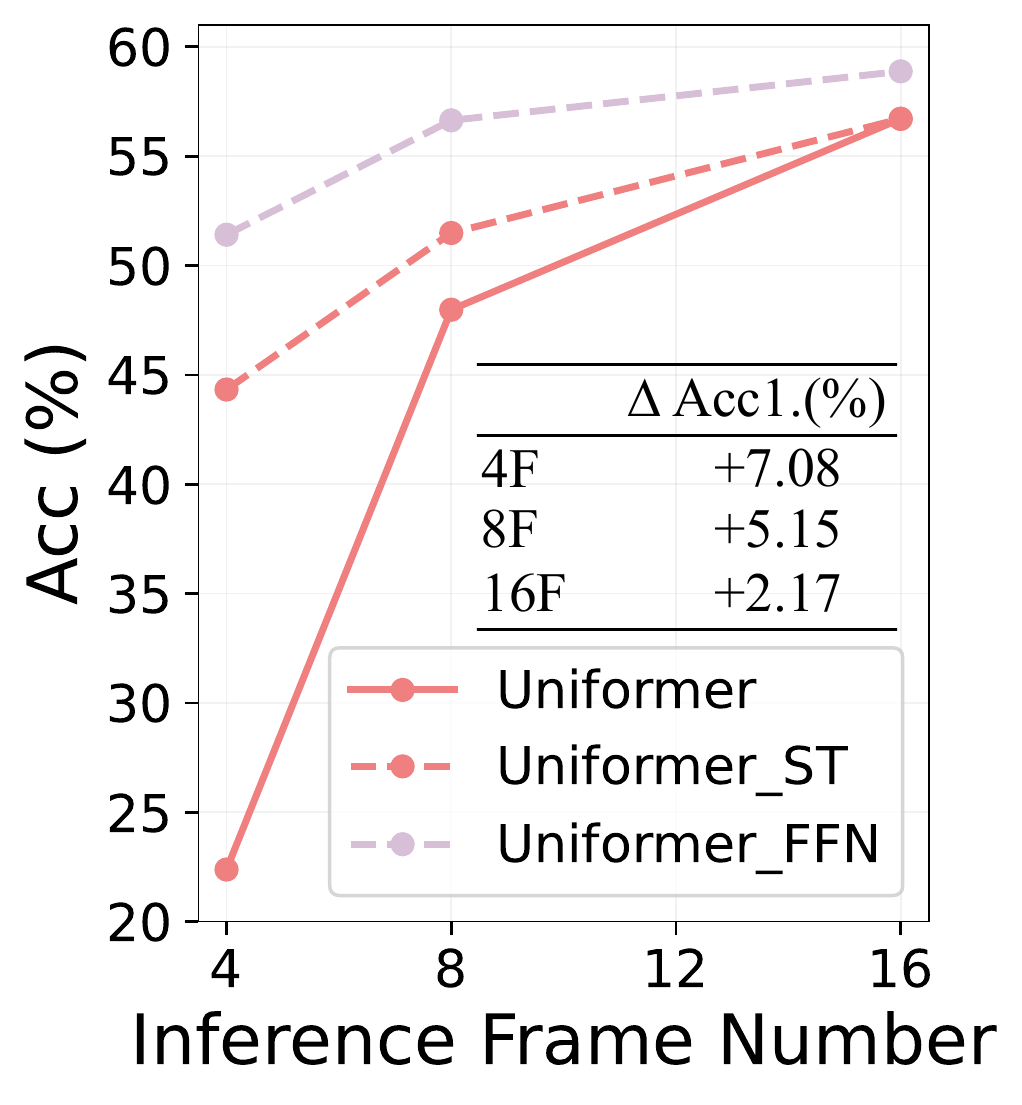}
          \vskip -0.05in
            \caption{Results on Uniformer.}
    \end{subfigure}
    \vskip -0.1in
    \caption{Validation results across different video recognition architectures on Something-Something V1 dataset, including 2D-network, 3D-network and Transformer-network. The improvements of FFN over ST are listed in the table.}
\label{fig:structure}
\vskip -0.05in
\end{figure*}

\begin{figure*}[t]
    \centering
    \begin{subfigure}[b]{0.32 \textwidth}
          \centering
          \includegraphics[width=\textwidth]{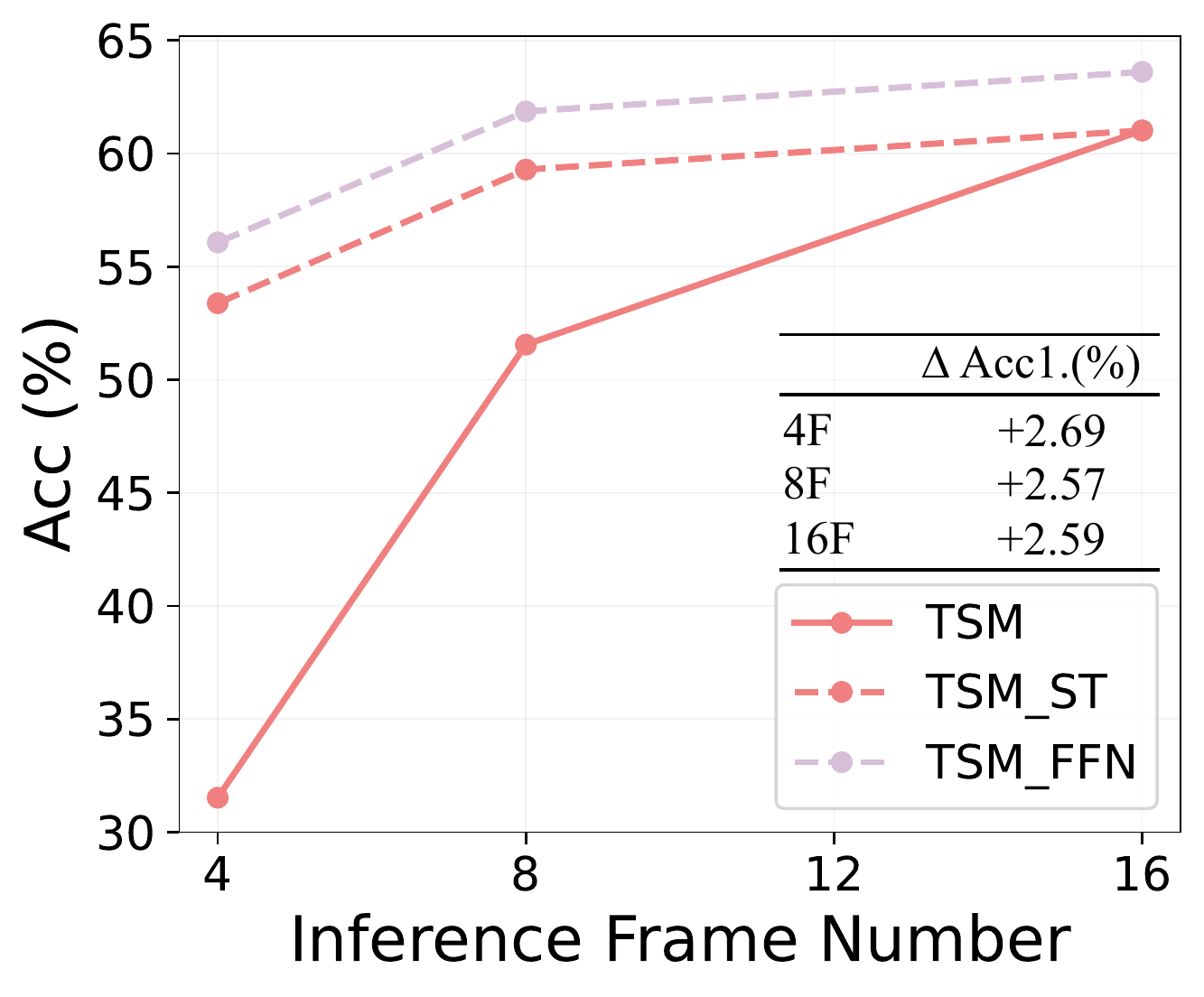}
          \vskip -0.05in
            \caption{Results on Something-Something V2 dataset.}
    \end{subfigure}
    \hfill
    \begin{subfigure}[b]{0.32 \textwidth}
            \centering
            \includegraphics[width=\textwidth]{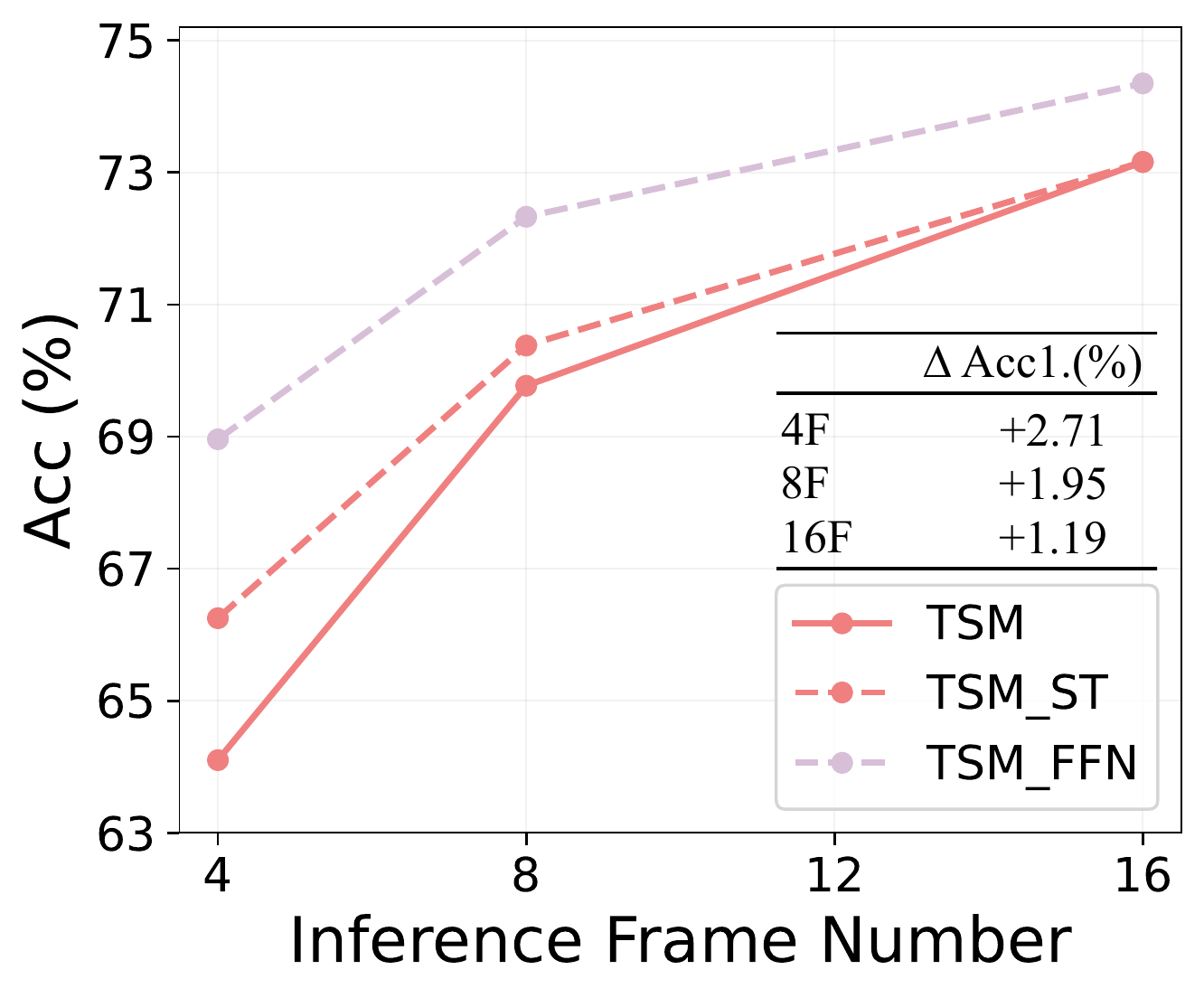}
          \vskip -0.05in
            \caption{Results on Kinetics400 dataset.}
    \end{subfigure}
    \hfill
    \begin{subfigure}[b]{0.32 \textwidth}
          \centering
          \includegraphics[width=\textwidth]{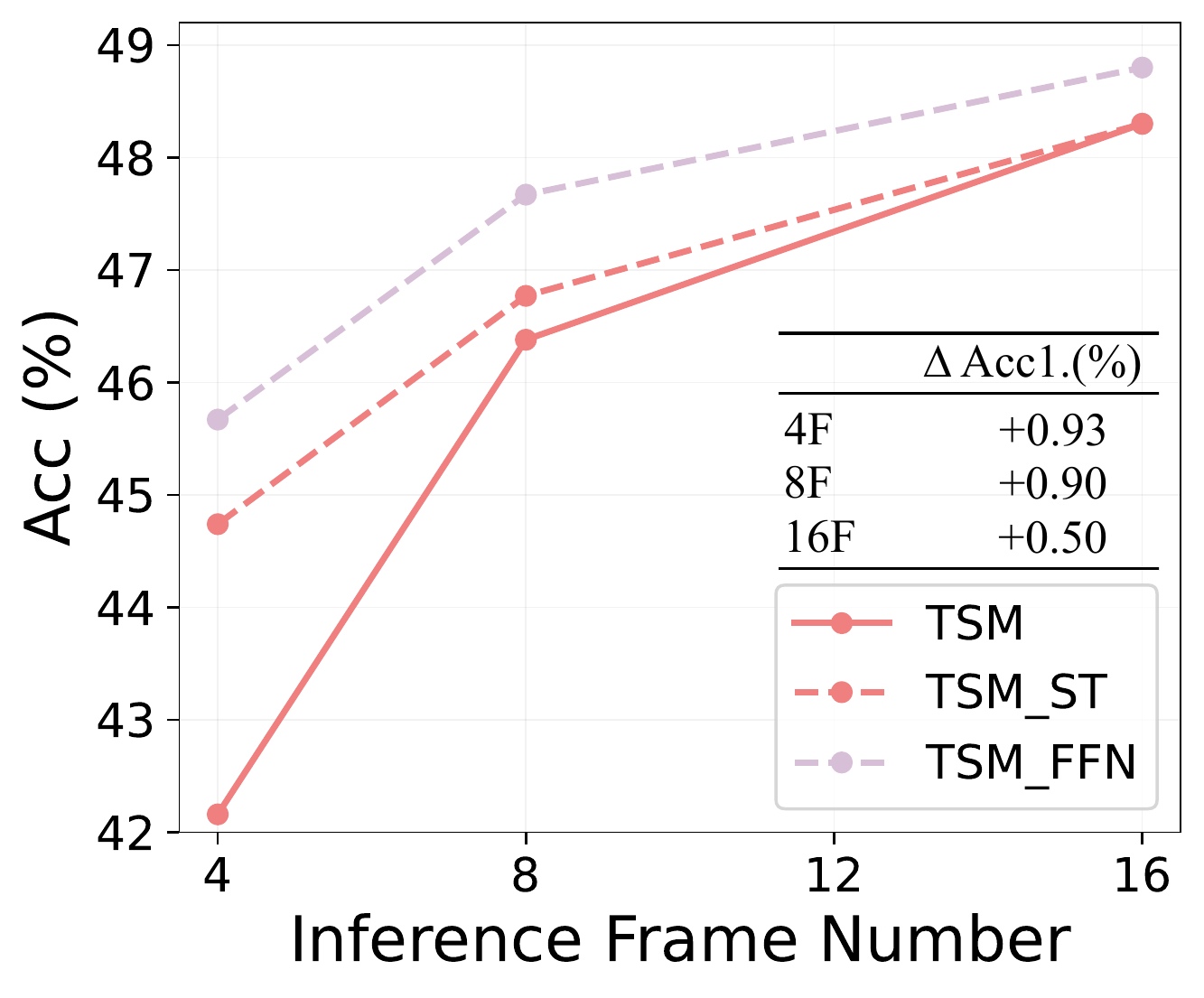}
          \vskip -0.05in
            \caption{Results on HMDB51 dataset.}
    \end{subfigure}
    \vskip -0.1in
    \caption{Validation results across various video recognition datasets. The improvements of FFN over ST are listed in the table.}
\label{fig:datasets}
\vskip -0.2in
\end{figure*}

\noindent\textbf{Datasets.} We conduct experiments on four datasets, including: (1) Something-Something V1$\&$V2~\cite{goyal2017something} include 98k and 194k videos, respectively. They contain strong temporal dependency and show the most significant Temporal Frequency Deviation phenomenon among all datasets. 
(2) Kinetics400~\cite{kay2017kinetics} is a large-scale dataset with 400 classes. 
(3) HMDB51~\cite{kuehne2011hmdb} is composed of 6,766 videos which can be categorized into 51 classes. We utilize the original three training/testing splits for training and evaluation.

\noindent\textbf{Implementation Details.} We uniformly sample 4/8/16 frames for $v^{L}$, $v^{M}$ and $v^{H}$ in all methods except for SlowFast~\cite{feichtenhofer2019slowfast} which samples 16/32/64 frames for fast pathway. For the baseline results, we train all methods with $v^{H}$ and evaluate them at $v^{L}$, $v^{M}$ and $v^{H}$. Separated Training (ST) denotes training the network at $v^{L}$, $v^{M}$, and $v^{H}$ individually and evaluating them at the frame used in training.

\noindent\textbf{Baseline Methods.} In addition to Separated Training (ST) introduced before, we provide four more baseline methods for this problem: 
(1) \textbf{Mixed Sampling}: We sample 4 and 16 frames for $v_{i}^{L}$ and $v_{i}^{H}$, respectively. Then we randomly choose 4 consecutive frames $v_{i}^{H'}$ from $v_{i}^{H}$ and apply mixup~\cite{zhang2017mixup} to integrate $v_{i}^{L}$ into $v_{i}^{H}$.
The hyperparameter $\rho$ decides the probability of whether to apply Mixed Sampling at each iteration. 
(2) \textbf{Proportional Sampling}: We let the network randomly sample 4 frames or 16 frames at each iteration as this pair has the most significant Temporal Frequency Deviation phenomenon. The hyperparameter $\varrho$ denotes the probability to sample 16 frames for every iteration.
(3) \textbf{Fine-tuning}: We first train the model at 16 Frame and then fine-tune it at 4 Frame.
(4) \textbf{Ensemble}: We make use of the models that are individually trained at 4,8 and 16 Frame and averagely ensemble them to form a new model.

\subsection{Main Results}

\noindent\textbf{Comparison with Baseline Methods.} Tab.~\ref{tab:ensemble} shows that Proportional Sampling and Mixed Sampling help to alleviate Temporal Frequency Deviation as the performance at Frame 4/8 is better than the inference results of the model trained with standard protocol. Nevertheless, the increase is obtained at the cost of an accuracy drop at Frame 16.
Then, we adjust the hyperparameter and the results show that both methods seem to provide a trade-off solution for this problem: if the performance at low frames is better, the results at high frame numbers will be worse. 

Fine-tuning helps the baseline method to exhibit comparable performance with ST at 4 Frame, but at a loss of forgetting the knowledge at 16 Frame. Ensemble outperforms ST at Frame 8 with the cost of multiplying computation. Besides, its performance at 4/16 Frame is worse than ST which means that it still cannot effectively resolve Temporal Frequency Deviation problem. While FFN shows stronger results compared to ST and Ensemble at all frames with negligible added computation. Moreover, compared to ST and Ensemble which 
need repetitive training operations and multiplying storage costs,
our method is trained for only one time, but can be evaluated at multiple frames, reducing the parameters of saving multiple models significantly which promises its applications on edge devices.

\noindent\textbf{Performance Analysis across Architectures.}\label{sec:structures} We further validate FFN on different architectures in Fig.~\ref{fig:structure}.
We first build our method on TSM~\cite{lin2019tsm} which does not contain any parameters in the temporal modeling module. FFN exhibits advantages in performance at all frames compared to baseline TSM
and ST. Then, we implement FFN on TEA~\cite{li2020tea} which involves convolutions and normalization in the temporal modeling module
and our results also surpass ST at all frames. Moreover, we extend FFN to 3D-network: SlowFast~\cite{feichtenhofer2019slowfast} and Transformer-network: Uniformer~\cite{li2022uniformer}. The results exhibit similar improvements at all frame numbers compared to ST which validates the flexibility and generalization ability of our method. Note that FFN introduces less than 1$\%$ extra computation but reduces the memory costs of storing individual models by multiple times.

\noindent\textbf{Performance Analysis across Datasets.} In this part, we empirically evaluate FFN on various datasets in Fig.~\ref{fig:datasets}, including Something-Something V2, Kinetics400 and HMDB51. The first observation is that Temporal Frequency Deviation phenomenon is less obvious on Kinetics400 and HMDB51 as these two datasets contain less temporal information. 
Nevertheless, FFN continuously improves the accuracy of ST on these datasets as well. For example, there are 2.71/1.95/1.19$\%$ performance gains at Frame 4/8/16 on Kinetics400 which further demonstrate the generalization ability of our design.

\subsection{Inference at Any Frame}\label{sec:anyframe}

We have proved that FFN can outperform ST at the frame numbers used in training, but the evaluation at other frames which are not included in training remains untouched. Motivated by Nearby Alleviation in Sec.~\ref{sec:near}, we provide a naive inference paradigm to enable FFN to be evaluated at any frame. Given a frame number $n$ at inference phase, we will calculate the frame difference with $L$, $M$ and $H$, and activate the sub-network with the minimal difference for validation. If the frame difference is the same for two sub-networks, we will choose the one which corresponds higher frame number by default. In this manner, we can evaluate at other frame numbers which are not used in training.

\begin{figure}[t]
\begin{center}
\scalebox{0.45}{\includegraphics[width=\textwidth]{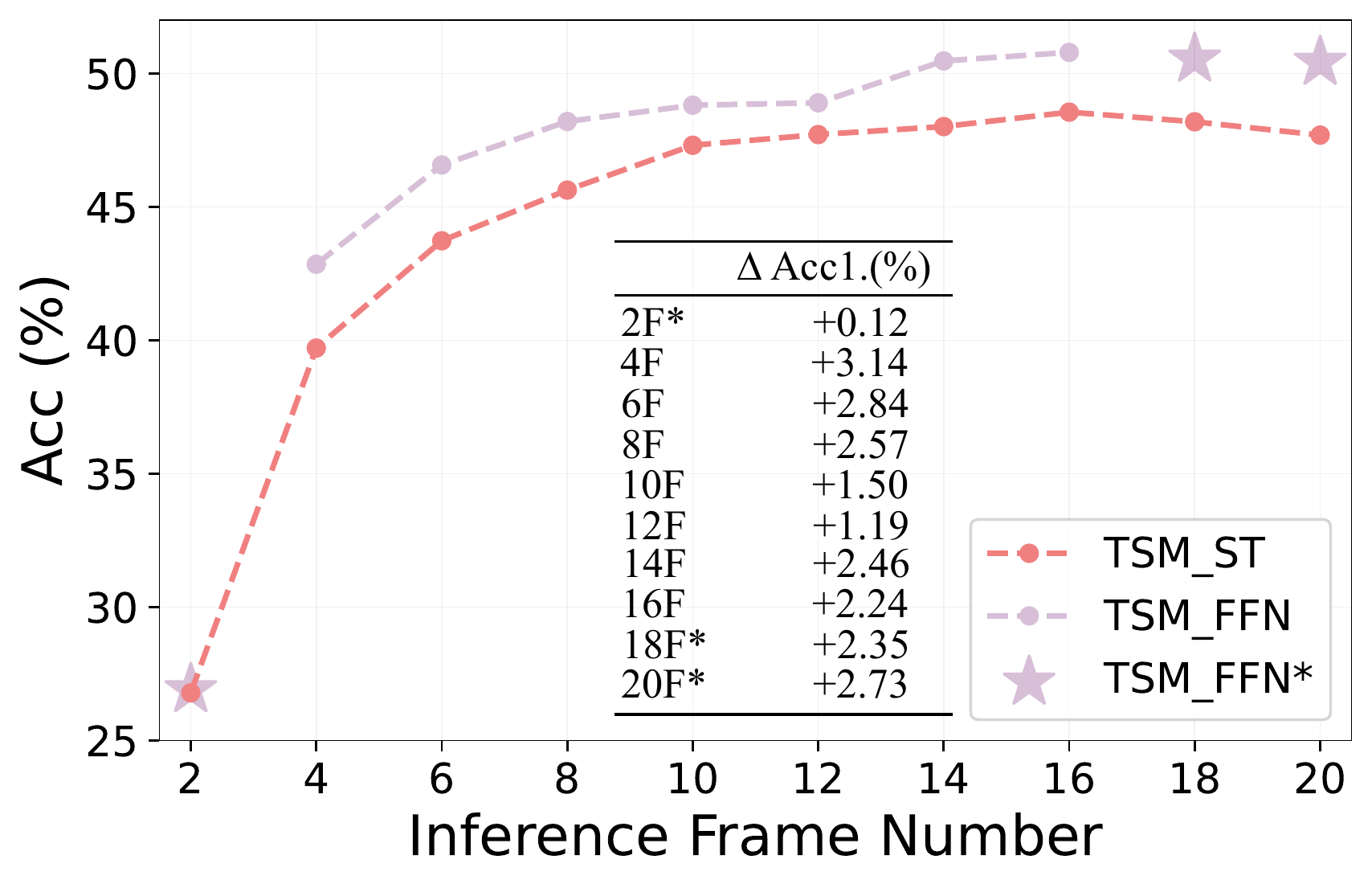}}
\end{center}
\vskip -0.3in
\caption{Validation results of FFN at various frame numbers on Something-Something V1. The improvements of FFN over ST are listed in the table. Outbound results are denoted with *.}
\label{fig:any}
\end{figure}

\noindent\textbf{Inbound Results.} Fig.~\ref{fig:any} shows that FFN outperforms ST at all frames within the range of 4-16 which are utilized in training. Though the improvement at 12 Frame is less obvious compared to other frames as it is the middle of 8/16 Frame which benefits the least from Nearby Alleviation.

\noindent\textbf{Outbound Results.} Moreover, we evaluate FFN at frames that are out of the bound of 4-16. One can observe that FFN even exhibits better performance compared to ST at Frame 2/18/20 which further demonstrates its generalization capacity at non-seen frames.

\subsection{Ablation}\label{sec:ablation}

\noindent\textbf{Input Sequences Combinations.}\label{sec:sequences} Shown in Tab.~\ref{tab:number}, we import different numbers of sequences to FFN and evaluate their performance at various frames. First, we can observe that FFN(2) outperforms ST at 4/16 Frame which are used in training, but its performance at 8/12 Frame is worse than ST because of the missing middle sequence in training. In contrast, both FFN(3) and FFN(4) obtain higher accuracy at all frames compared to ST which can be attributed to the utilization of the middle sequence in training so that the Temporal Frequency Deviation at nearby frames can be mitigated by Nearby Alleviation. FFN(4) obtains better results at Frame 4/8/12 because of the added sequence, but it will cost more time and resources during training.

\begin{table}[t]
\centering
\vskip -0.05in
\scalebox{0.85}{\begin{tabular}{lccccccc}
\toprule
\multirow{2}*{Method} & \multirow{2}*{Sequences} & \multicolumn{4}{c}{Top-1 Acc. ($\%$)} \\
\cmidrule(lr){3-7}
& & 4F & 8F & 12F & 16F \\
\midrule
TSM-ST~\cite{lin2019tsm}  & - & 39.71 & 45.63 & 47.71 & 48.55 \\
\hdashline
TSM-FFN(2)  & 4/16 & 41.69 & 37.93 & 48.10 & 49.79 \\
TSM-FFN(3)  & 4/8/16 & 42.85 & 48.20 & 48.90 & \textbf{50.79} \\
TSM-FFN(4)  & 4/8/12/16 & \textbf{43.40} & \textbf{48.66} & \textbf{49.77} & 50.63 \\
\bottomrule
\end{tabular}}
\vskip -0.05in
\caption{Experiments of different input sequences combinations on Something-Something V1. The best results are bold-faced.}
\label{tab:number}
\vskip -0.2in
\end{table}

\noindent\textbf{Frame Numbers.} 
We sample more frames to $v^{H}$ in this section and import four sequences with 4/8/16/24 frames to FFN, respectively. The first observation from Tab.~\ref{tab:more_frames} is that the performance of TSM-ST (24F) is even a little bit lower than TSM-ST (16F) which can be attributed to relatively simple temporal modeling measure of TSM. However, FFN still obtains better performance compared with ST at all frames and achieves the highest accuracy at 24 Frame, owing to the design of Temporal Distillation.

\noindent\textbf{Design Choices.} We conduct ablation to verify the effectiveness of our designs in Tab.~\ref{tab:ablation}. First, we build FFN with shared normalization and one can observe an obvious performance drop at 4/16 Frame due to the shift in normalization statistics.
Then, we remove Weight Alteration in the convolution block and it exhibits worse performance at all frames
which proves the strength of Multi-Frequency Adaptation (MFAD) as it increases the representation abilities of the sub-networks at corresponding frames.
Further, we optimize FFN by calculating CE loss on the predictions of all sub-networks respectively and do not utilize KL divergence loss for optimization. The results are clearly worse than ST which suggests that Temporal Distillation is a necessary component for Multi-Frequency Alignment (MFAL). Finally, we specialize convolutions (w/o WS) and note that this operation will multiply the parameters by three times. We can observe both FFN and FFN(w/o WA) outperform specialized convolutions at all frames with fewer parameters which demonstrates the effectiveness of MFAL to learn 
temporal frequency invariant representations.

\begin{table}[t]
\centering
\scalebox{0.8}{\begin{tabular}{lcccccc}
\toprule
\multirow{2}*{Method} & \multirow{2}*{Parameters} & \multicolumn{4}{c}{Top-1 Acc. ($\%$)} \\
\cmidrule(lr){3-7}
& & 4F & 8F & 16F & 24F \\
\midrule
TSM-ST~\cite{lin2019tsm}  & 25.6$\times$4M & 39.71 & 45.63 & 48.55 & 47.90 \\
\hdashline
TSM-FFN & 25.7M & \textbf{41.28} & \textbf{46.72} & \textbf{49.79} &  \textbf{49.95} \\
\bottomrule
\end{tabular}}
\vskip -0.05in
\caption{Validation results of FFN with more sampled frames on Something-Something V1. The best results are bold-faced.}
\label{tab:more_frames}
\end{table}

\begin{table}[t]
\centering
\vskip -0.05in
\scalebox{0.8}{\begin{tabular}{lcccccc}
\toprule
\multirow{2}*{Method} & \multirow{2}*{Parameters} & \multirow{2}*{Specification} & \multicolumn{3}{c}{Top-1 Acc. ($\%$)} \\
\cmidrule(lr){4-7}
& & & 4F & 8F & 16F \\
\midrule
TSM-ST~\cite{lin2019tsm}  & 25.6$\times$3M & - & 39.71 & 45.63 & 48.55 \\
\hdashline
TSM-FFN & 25.7M & w/o SN & 35.79 & 46.80 & 44.62 \\
TSM-FFN & 25.6M & w/o WA & 41.91 & 47.92 & 49.84 \\
TSM-FFN & 25.7M & w/o TD & 39.61 & 45.65 & 48.07 \\
TSM-FFN & 25.6$\times$3M & w/o WS & 41.51 & 47.16 & 48.23 \\
\midrule
TSM-FFN & 25.7M & - & \textbf{42.85} & \textbf{48.20} & \textbf{50.79} \\
\bottomrule
\end{tabular}}
\vskip -0.05in
\caption{Ablation of design choices of FFN on Something-Something V1. SN, TD, WA, WS denotes Specialized Normalization, Temporal Distillation, Weight Alteration, Weight Sharing, respectively. The best results are bold-faced.}
\label{tab:ablation}
\vskip -0.2in
\end{table}

\section{Conclusion and Limitations}

In this paper, we reveal Temporal Frequency Deviation phenomenon and propose Frame Flexible Network (FFN) to address it. 
Specifically, we propose Multi-Frequency Alignment to learn temporal frequency invariant representations and present Multi-Frequency Adaptation to further strengthen the representation ability.
Extensive experiments demonstrate that FFN, which only requires one-shot training, can be evaluated at multiple frames and outperforms Separated Training with significantly fewer parameters, making it favorable for applications on edge devices.

One limitation of FFN is that, it requires more GPU memory during training as we import several input sequences. Second, FFN introduces slightly extra computation because of Weight Alteration. In future work, we are interested in improving the training efficiency of FFN.

\vspace{-8pt}
\section*{Acknowledgment}
\vspace{-6pt}
Research was sponsored by the DEVCOM Analysis Center and was accomplished under Cooperative Agreement Number W911NF-22-2-0001. The views and conclusions contained in this document are those of the authors and should not be interpreted as representing the official policies, either expressed or implied, of the Army Research Office or the U.S. Government. The U.S. Government is authorized to reproduce and distribute reprints for Government purposes notwithstanding any copyright notation herein.

{\small
\bibliographystyle{ieee_fullname}
\bibliography{egbib}

\begin{thebibliography}{10}\itemsep=-1pt

\bibitem{ba2016layer}
Jimmy~Lei Ba, Jamie~Ryan Kiros, and Geoffrey~E Hinton.
\newblock Layer normalization.
\newblock {\em arXiv preprint arXiv:1607.06450}, 2016.

\bibitem{carreira2017quo}
Joao Carreira and Andrew Zisserman.
\newblock Quo vadis, action recognition? a new model and the kinetics dataset.
\newblock In {\em CVPR}, 2017.

\bibitem{chollet2017xception}
Fran{\c{c}}ois Chollet.
\newblock Xception: Deep learning with depthwise separable convolutions.
\newblock In {\em CVPR}, 2017.

\bibitem{dosovitskiy2020image}
Alexey Dosovitskiy, Lucas Beyer, Alexander Kolesnikov, Dirk Weissenborn,
  Xiaohua Zhai, Thomas Unterthiner, Mostafa Dehghani, Matthias Minderer, Georg
  Heigold, Sylvain Gelly, et~al.
\newblock An image is worth 16x16 words: Transformers for image recognition at
  scale.
\newblock {\em arXiv preprint arXiv:2010.11929}, 2020.

\bibitem{fan2021multiscale}
Haoqi Fan, Bo Xiong, Karttikeya Mangalam, Yanghao Li, Zhicheng Yan, Jitendra
  Malik, and Christoph Feichtenhofer.
\newblock Multiscale vision transformers.
\newblock In {\em ICCV}, 2021.

\bibitem{feichtenhofer2019slowfast}
Christoph Feichtenhofer, Haoqi Fan, Jitendra Malik, and Kaiming He.
\newblock Slowfast networks for video recognition.
\newblock In {\em ICCV}, 2019.

\bibitem{goyal2017something}
Raghav Goyal, Samira Ebrahimi~Kahou, Vincent Michalski, Joanna Materzynska,
  Susanne Westphal, Heuna Kim, Valentin Haenel, Ingo Fruend, Peter Yianilos,
  Moritz Mueller-Freitag, et~al.
\newblock The" something something" video database for learning and evaluating
  visual common sense.
\newblock In {\em ICCV}, 2017.

\bibitem{he2016deep}
Kaiming He, Xiangyu Zhang, Shaoqing Ren, and Jian Sun.
\newblock Deep residual learning for image recognition.
\newblock In {\em CVPR}, 2016.

\bibitem{howard2017mobilenets}
Andrew~G Howard, Menglong Zhu, Bo Chen, Dmitry Kalenichenko, Weijun Wang,
  Tobias Weyand, Marco Andreetto, and Hartwig Adam.
\newblock Mobilenets: Efficient convolutional neural networks for mobile vision
  applications.
\newblock {\em arXiv preprint arXiv:1704.04861}, 2017.

\bibitem{huang2017multi}
Gao Huang, Danlu Chen, Tianhong Li, Felix Wu, Laurens Van Der~Maaten, and
  Kilian~Q Weinberger.
\newblock Multi-scale dense networks for resource efficient image
  classification.
\newblock {\em arXiv preprint arXiv:1703.09844}, 2017.

\bibitem{kay2017kinetics}
Will Kay, Joao Carreira, Karen Simonyan, Brian Zhang, Chloe Hillier, Sudheendra
  Vijayanarasimhan, Fabio Viola, Tim Green, Trevor Back, Paul Natsev, et~al.
\newblock The kinetics human action video dataset.
\newblock {\em arXiv preprint arXiv:1705.06950}, 2017.

\bibitem{korbar2019scsampler}
Bruno Korbar, Du Tran, and Lorenzo Torresani.
\newblock Scsampler: Sampling salient clips from video for efficient action
  recognition.
\newblock In {\em ICCV}, 2019.

\bibitem{kuehne2011hmdb}
Hildegard Kuehne, Hueihan Jhuang, Est{\'\i}baliz Garrote, Tomaso Poggio, and
  Thomas Serre.
\newblock Hmdb: a large video database for human motion recognition.
\newblock In {\em ICCV}, 2011.

\bibitem{kullback1997information}
Solomon Kullback.
\newblock {\em Information theory and statistics}.
\newblock Courier Corporation, 1997.

\bibitem{li2020learning}
Duo Li, Anbang Yao, and Qifeng Chen.
\newblock Learning to learn parameterized classification networks for scalable
  input images.
\newblock In {\em ECCV}, 2020.

\bibitem{li2022uniformer}
Kunchang Li, Yali Wang, Junhao Zhang, Peng Gao, Guanglu Song, Yu Liu, Hongsheng
  Li, and Yu Qiao.
\newblock Uniformer: Unifying convolution and self-attention for visual
  recognition.
\newblock {\em arXiv preprint arXiv:2201.09450}, 2022.

\bibitem{li2020tea}
Yan Li, Bin Ji, Xintian Shi, Jianguo Zhang, Bin Kang, and Limin Wang.
\newblock Tea: Temporal excitation and aggregation for action recognition.
\newblock In {\em CVPR}, 2020.

\bibitem{lin2019tsm}
Ji Lin, Chuang Gan, and Song Han.
\newblock Tsm: Temporal shift module for efficient video understanding.
\newblock In {\em ICCV}, 2019.

\bibitem{liu2021swin}
Ze Liu, Yutong Lin, Yue Cao, Han Hu, Yixuan Wei, Zheng Zhang, Stephen Lin, and
  Baining Guo.
\newblock Swin transformer: Hierarchical vision transformer using shifted
  windows.
\newblock In {\em ICCV}, 2021.

\bibitem{liu2022video}
Ze Liu, Jia Ning, Yue Cao, Yixuan Wei, Zheng Zhang, Stephen Lin, and Han Hu.
\newblock Video swin transformer.
\newblock In {\em CVPR}, 2022.

\bibitem{meng2020ar}
Yue Meng, Chung-Ching Lin, Rameswar Panda, Prasanna Sattigeri, Leonid
  Karlinsky, Aude Oliva, Kate Saenko, and Rogerio Feris.
\newblock Ar-net: Adaptive frame resolution for efficient action recognition.
\newblock In {\em ECCV}, 2020.

\bibitem{pan2022st}
Junting Pan, Ziyi Lin, Xiatian Zhu, Jing Shao, and Hongsheng Li.
\newblock St-adapter: Parameter-efficient image-to-video transfer learning for
  action recognition.
\newblock {\em arXiv preprint arXiv:2206.13559}, 2022.

\bibitem{pfeiffer2020adapterfusion}
Jonas Pfeiffer, Aishwarya Kamath, Andreas R{\"u}ckl{\'e}, Kyunghyun Cho, and
  Iryna Gurevych.
\newblock Adapterfusion: Non-destructive task composition for transfer
  learning.
\newblock {\em arXiv preprint arXiv:2005.00247}, 2020.

\bibitem{pfeiffer2020adapterhub}
Jonas Pfeiffer, Andreas R{\"u}ckl{\'e}, Clifton Poth, Aishwarya Kamath, Ivan
  Vuli{\'c}, Sebastian Ruder, Kyunghyun Cho, and Iryna Gurevych.
\newblock Adapterhub: A framework for adapting transformers.
\newblock {\em arXiv preprint arXiv:2007.07779}, 2020.

\bibitem{singh2021semi}
Ankit Singh, Omprakash Chakraborty, Ashutosh Varshney, Rameswar Panda, Rogerio
  Feris, Kate Saenko, and Abir Das.
\newblock Semi-supervised action recognition with temporal contrastive
  learning.
\newblock In {\em CVPR}, 2021.

\bibitem{sung2022vl}
Yi-Lin Sung, Jaemin Cho, and Mohit Bansal.
\newblock Vl-adapter: Parameter-efficient transfer learning for
  vision-and-language tasks.
\newblock In {\em CVPR}, 2022.

\bibitem{touvron2019fixing}
Hugo Touvron, Andrea Vedaldi, Matthijs Douze, and Herv{\'e} J{\'e}gou.
\newblock Fixing the train-test resolution discrepancy.
\newblock {\em NeurIPS}, 2019.

\bibitem{tran2015learning}
Du Tran, Lubomir Bourdev, Rob Fergus, Lorenzo Torresani, and Manohar Paluri.
\newblock Learning spatiotemporal features with 3d convolutional networks.
\newblock In {\em ICCV}, 2015.

\bibitem{wang2021tdn}
Limin Wang, Zhan Tong, Bin Ji, and Gangshan Wu.
\newblock Tdn: Temporal difference networks for efficient action recognition.
\newblock In {\em CVPR}, 2021.

\bibitem{wang2016temporal}
Limin Wang, Yuanjun Xiong, Zhe Wang, Yu Qiao, Dahua Lin, Xiaoou Tang, and Luc
  Van~Gool.
\newblock Temporal segment networks: Towards good practices for deep action
  recognition.
\newblock In {\em ECCV}, 2016.

\bibitem{wang2021adaptive}
Yulin Wang, Zhaoxi Chen, Haojun Jiang, Shiji Song, Yizeng Han, and Gao Huang.
\newblock Adaptive focus for efficient video recognition.
\newblock {\em arXiv preprint arXiv:2105.03245}, 2021.

\bibitem{wu2020dynamic}
Zuxuan Wu, Hengduo Li, Caiming Xiong, Yu-Gang Jiang, and Larry~Steven Davis.
\newblock A dynamic frame selection framework for fast video recognition.
\newblock {\em IEEE Transactions on Pattern Analysis and Machine Intelligence},
  2020.

\bibitem{yang2020video}
Ceyuan Yang, Yinghao Xu, Bo Dai, and Bolei Zhou.
\newblock Video representation learning with visual tempo consistency.
\newblock {\em arXiv preprint arXiv:2006.15489}, 2020.

\bibitem{yang2020mutualnet}
Taojiannan Yang, Sijie Zhu, Chen Chen, Shen Yan, Mi Zhang, and Andrew Willis.
\newblock Mutualnet: Adaptive convnet via mutual learning from network width
  and resolution.
\newblock In {\em ECCV}, 2020.

\bibitem{yang2021mutualnet}
Taojiannan Yang, Sijie Zhu, Matias Mendieta, Pu Wang, Ravikumar Balakrishnan,
  Minwoo Lee, Tao Han, Mubarak Shah, and Chen Chen.
\newblock Mutualnet: Adaptive convnet via mutual learning from different model
  configurations.
\newblock {\em IEEE Transactions on Pattern Analysis and Machine Intelligence},
  2021.

\bibitem{yu2019universally}
Jiahui Yu and Thomas~S Huang.
\newblock Universally slimmable networks and improved training techniques.
\newblock In {\em ICCV}, 2019.

\bibitem{yu2018slim}
Jiahui Yu, Linjie Yang, Ning Xu, Jianchao Yang, and Thomas Huang.
\newblock Slimmable neural networks.
\newblock {\em arXiv preprint arXiv:1812.08928}, 2018.

\bibitem{zhang2017mixup}
Hongyi Zhang, Moustapha Cisse, Yann~N Dauphin, and David Lopez-Paz.
\newblock mixup: Beyond empirical risk minimization.
\newblock {\em arXiv preprint arXiv:1710.09412}, 2017.

\bibitem{zhang2022minivit}
Jinnian Zhang, Houwen Peng, Kan Wu, Mengchen Liu, Bin Xiao, Jianlong Fu, and Lu
  Yuan.
\newblock Minivit: Compressing vision transformers with weight multiplexing.
\newblock In {\em CVPR}, 2022.

\bibitem{zhang2018shufflenet}
Xiangyu Zhang, Xinyu Zhou, Mengxiao Lin, and Jian Sun.
\newblock Shufflenet: An extremely efficient convolutional neural network for
  mobile devices.
\newblock In {\em CVPR}, 2018.

\bibitem{zhang2022look}
Yitian Zhang, Yue Bai, Huan Wang, Yi Xu, and Yun Fu.
\newblock Look more but care less in video recognition.
\newblock {\em arXiv preprint arXiv:2211.09992}, 2022.

\end{thebibliography}
}


\clearpage
\section*{Supplementary Material}

\subsection*{A.\quad Implementation Details}

The training data is randomly cropped to 224 $\times$ 224 and we perform random flipping except for Something-Something datasets. At inference stage, all frames will be center-cropped to 224 $\times$ 224 except SlowFast~\cite{feichtenhofer2019slowfast} which adopts the resolution of 256 $\times$ 256 for evaluation. We use one-clip one-crop per video during evaluation except Uniformer~\cite{li2022uniformer} which utilizes one-clip three-crop evaluation protocol. We train all models on NVIDIA Tesla V100 GPUs and adopt the same training hyperparameters with the official implementations.

\subsection*{B.\quad Results of Different Depths}

\begin{table}[h]
\centering
\vskip -0.1in
\caption{Experiments with different depths on Something-Something V1. The best results are bold-faced.}
\vskip -0.05in
\scalebox{0.85}{\begin{tabular}{lcccc}
\toprule
\multirow{2}*{Method} & \multicolumn{3}{c}{Top-1 Acc.($\%$)} \\
\cmidrule(lr){2-5}
& $v^{L}$ & $v^{M}$ & $v^{H}$ \\
\midrule
TSM(R18)~\cite{lin2019tsm}  & 16.82 & 33.12 & 42.95 \\
TSM(R18)-ST  & 32.33 & 38.21 & 42.95 \\
TSM(R18)-FFN & \textbf{36.83}(4.50$\uparrow$) & \textbf{41.61}(3.40$\uparrow$) & \textbf{43.57}(0.62$\uparrow$) \\
\hdashline
TSM(R101)~\cite{lin2019tsm}  & 22.15 & 39.30 & 49.57 \\
TSM(R101)-ST  & 40.76 & 46.96 & 49.57 \\
TSM(R101)-FFN & \textbf{45.15}(4.39$\uparrow$) & \textbf{50.24}(3.28$\uparrow$) & \textbf{51.79}(2.22$\uparrow$) \\
\bottomrule
\end{tabular}}
\label{tab:depths}
\end{table}

As we have shown in the main text, Temporal Frequency Deviation phenomenon exists in different depths of the network which means it has no relation to the representation ability. But whether FFN can address this issue at other depths remains a problem. As previous experiments are built on ResNet-50~\cite{he2016deep}, we conduct experiments on ResNet-18, ResNet-101 and include their results in Tab.~\ref{tab:depths}. The results show that FFN outperforms Separated Training (ST) at different frame numbers which proves that FFN can effectively resolve Temporal Frequency Deviation problem regardless of the depths of the deep network.

\subsection*{C.\quad Results of Different Middle Sequences}

\begin{table*}[t]
\centering
\caption{Experiments with different middle sequences on Something-Something V1. The best results are bold-faced.}
\scalebox{1}{\begin{tabular}{lccccccccc}
\toprule
\multirow{2}*{Method} & \multirow{2}*{$v^{M}$} & \multicolumn{7}{c}{Top-1 Acc.($\%$)} \\
\cmidrule(lr){3-10}
& & 4 Frame & 6 Frame & 8 Frame & 10 Frame & 12 Frame & 14 Frame & 16 Frame \\
\midrule
TSM~\cite{lin2019tsm} & - & 20.60 & 30.23 & 37.36 & 42.72 & 45.97 & 47.49 & 48.55 \\
TSM-ST  & - & 39.71 & 43.73 & 45.63 & 47.31 & 47.71 & 48.01 & 48.55 \\
\hdashline
TSM-FFN  & 8F & 42.85 & \textbf{46.57} & \textbf{48.20} & 48.81 & 48.90 & 50.47 & 50.79 \\
TSM-FFN  & 10F & \textbf{43.10} & 44.77 & 47.81 & \textbf{49.26} & 49.63 & \textbf{50.67} & \textbf{51.12} \\
TSM-FFN  & 12F & 42.92 & 43.57 & 46.82 & 48.85 & \textbf{49.73} & 50.40 & 50.79 \\
\bottomrule
\end{tabular}}
\label{tab:middle}
\end{table*}

Another design choice in our method is the selection of middle sequence $v^{M}$, as $v^{L}$ and $v^{H}$ are usually set at first based on the range of the computations. Thus, we sample 8/10/12 frames for $v^{M}$ respectively and evaluate them at various frame numbers in Tab.~\ref{tab:middle}. When we sample 8 frames for $v^{M}$, FFN obtains the best performance at 8 Frame compared to the other two choices and the phenomenon is the same when sampling 10 or 12 frames for $v^{M}$. This meets our expectation as the specialized normalization for $v^{M}$ learns its corresponding transformation. Overall, all three choices lead to consistent improvement over Separated Training (ST) at all frames.

\subsection*{D.\quad Any Frame Inference of Input Sequences Combinations}

\begin{table*}[t]
\centering
\caption{Any frame inference results of input sequences combinations on Something-Something V1. The best results are bold-faced.}
\scalebox{1}{\begin{tabular}{lccccccccc}
\toprule
\multirow{2}*{Method} & \multirow{2}*{Sequences} & \multicolumn{7}{c}{Top-1 Acc.($\%$)} \\
\cmidrule(lr){3-10}
& & 4 Frame & 6 Frame & 8 Frame & 10 Frame & 12 Frame & 14 Frame & 16 Frame \\
\midrule
TSM~\cite{lin2019tsm} & - & 20.60 & 30.23 & 37.36 & 42.72 & 45.97 & 47.49 & 48.55 \\
TSM-ST  & - & 39.71 & 43.73 & 45.63 & 47.31 & 47.71 & 48.01 & 48.55 \\
\hdashline
TSM-FFN(2)  & 4/16 & 41.69 & 42.07 & 37.93 & 46.11 & 48.10 & 49.37 & 49.79 \\
TSM-FFN(3)  & 4/8/16 & 42.85 & \textbf{46.57} & 48.20 & 48.81 & 48.90 & \textbf{50.47} & \textbf{50.79} \\
TSM-FFN(4)  & 4/8/12/16 & \textbf{43.40} & 46.51 & \textbf{48.66} & \textbf{48.92} & \textbf{49.77} & 50.11 & 50.63 \\
\bottomrule
\end{tabular}}
\label{tab:sequences}
\vskip -0.1in
\end{table*}

In the main text, we have conducted the ablation of input sequences combinations. We further validate the three models at more fine-grained frame numbers with the proposed inference paradigm and the results are shown in Tab.~\ref{tab:sequences}. One can observe that FFN(2) obtains lower accuracy compared to ST at 6/8/10 Frame because of the missing middle sequence. While FFN(4) achieves the highest performance at 8/10/12 Frame as the introduced sequence at Frame 12 will alleviate the Temporal Frequency Deviation nearby.

\subsection*{E.\quad Further Verification of Nearby Alleviation}

\begin{figure}[h]
\begin{center}
\vskip -0.1in
\scalebox{0.35}{\includegraphics[width=\textwidth]{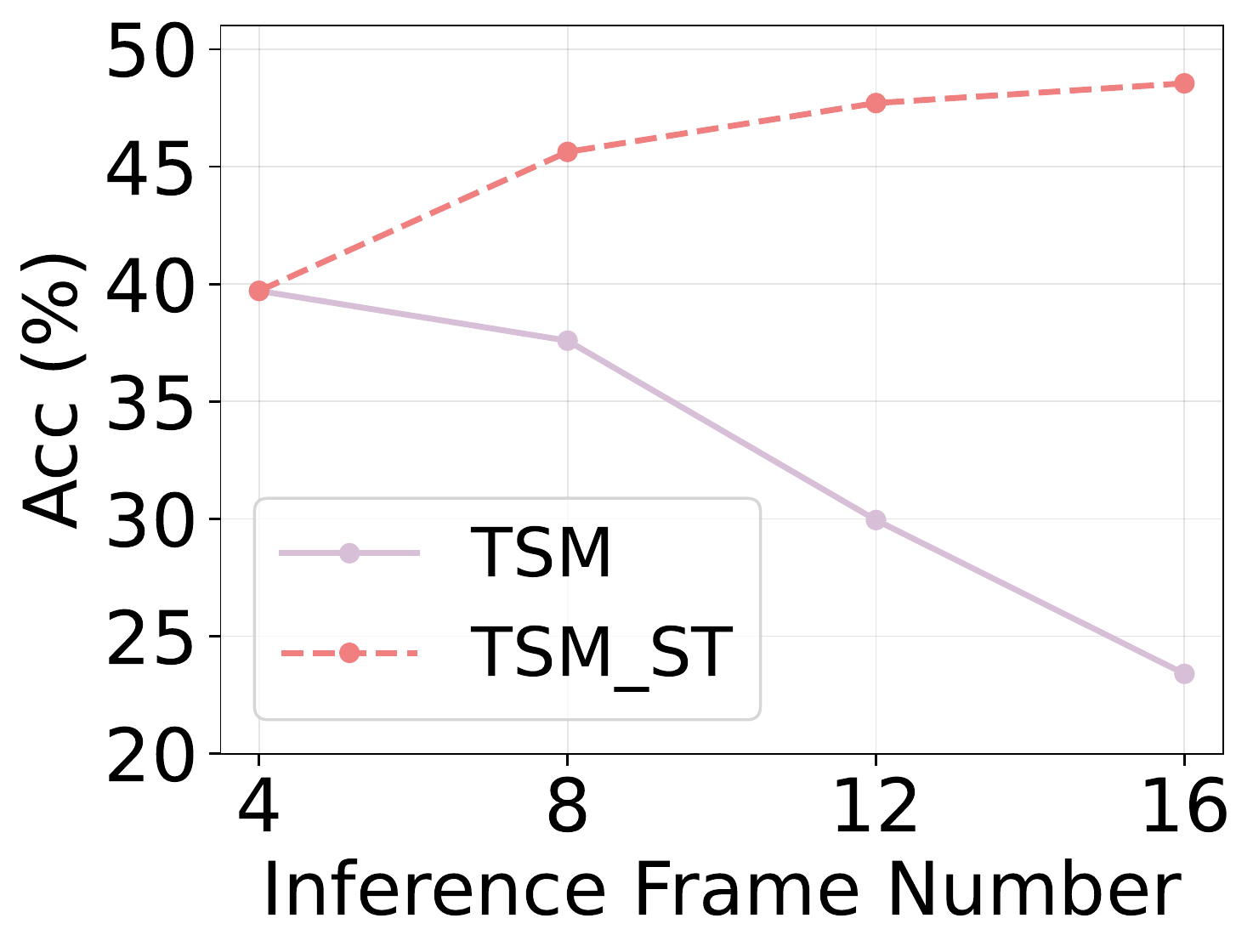}}
\end{center}
\vskip -0.2in
\caption{Validation results of TSM which is trained at 4 Frame on Something-Something V1 dataset.}
\label{fig:further_near}
\vskip -0.1in
\end{figure}

In previous parts, we have conducted experiments which train the model at Frame 8/12/16 and evaluate their performance at different frames. Here we further train the model at 4 Frame and show the validation results in Fig.~\ref{fig:further_near}. Similarly, we can observe that frames close to 4 exhibit the slightest performance drop as their normalization statistics is more similar with frame 4 which further verifies the Nearby Alleviation phenomenon.

\subsection*{F.\quad Statistics of Normalization Shifting}

\begin{figure}[h]
\vskip -0.1in
    \centering
    \begin{subfigure}[b]{0.23 \textwidth}
           \centering
           \includegraphics[width=\textwidth]{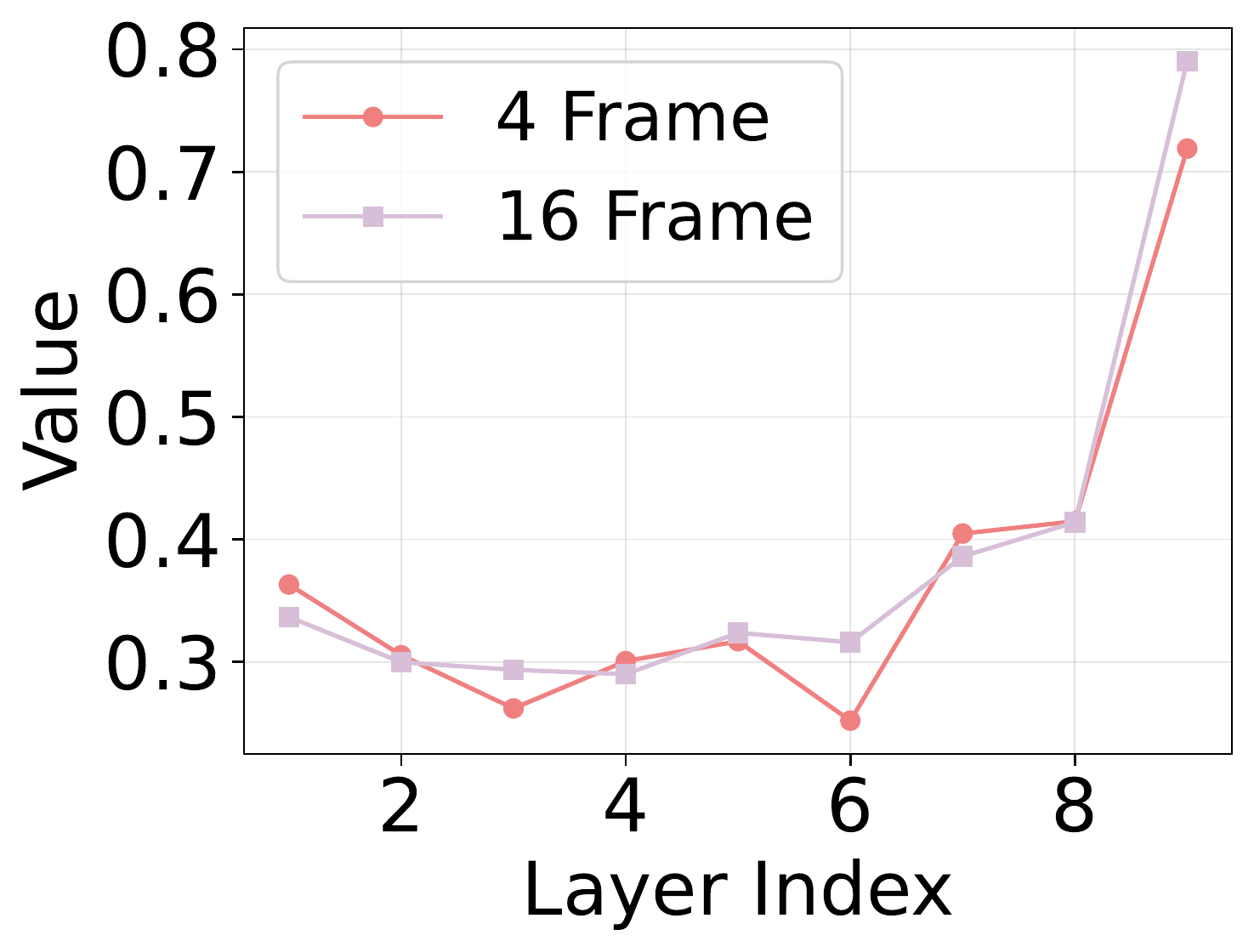}
           \vskip -0.05in
            \caption{Scale: $\gamma$}
    \end{subfigure}
    \hfill
    \begin{subfigure}[b]{0.23 \textwidth}
            \centering
            \includegraphics[width=\textwidth]{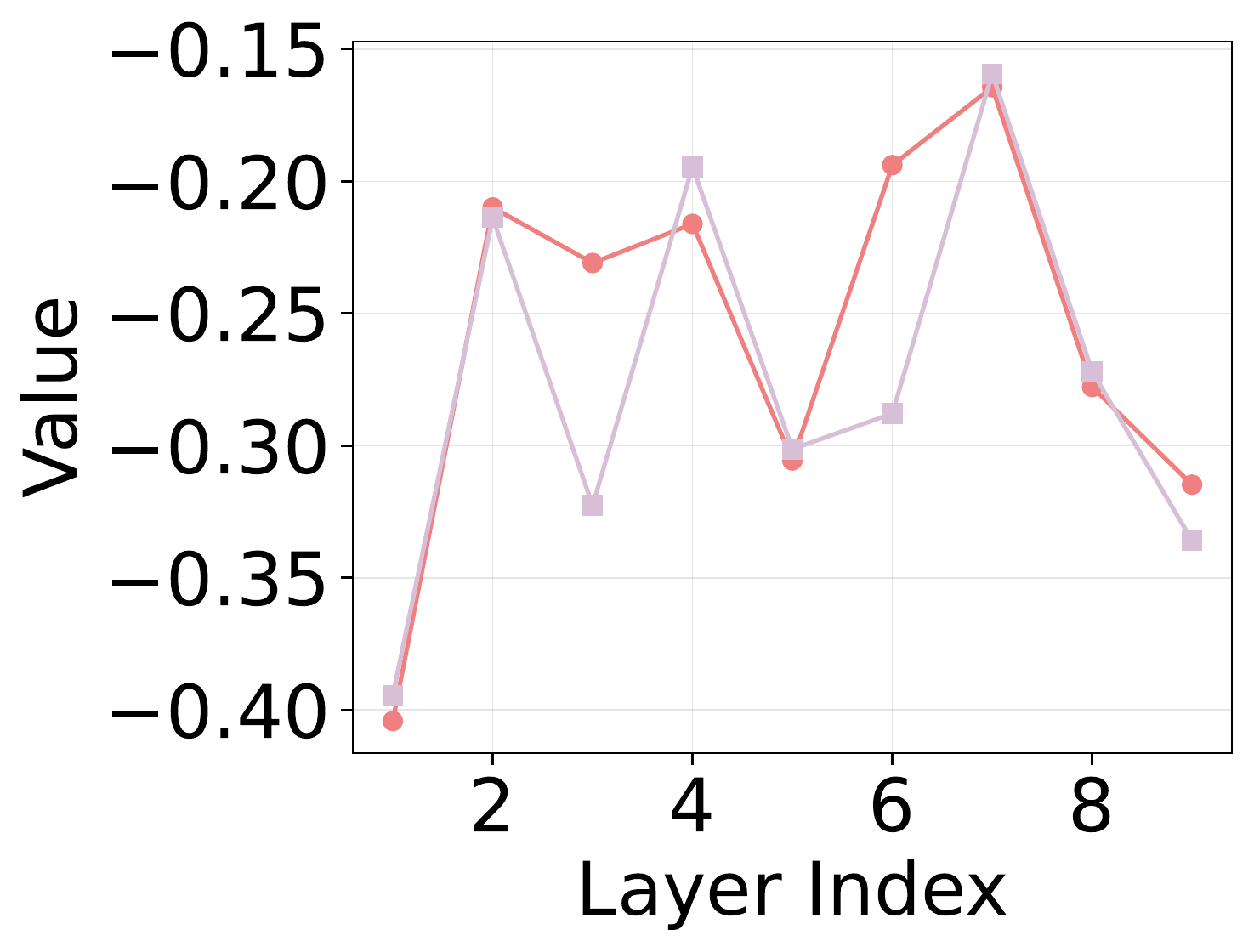}
           \vskip -0.05in
            \caption{Bias: $\beta$}
    \end{subfigure}
    \vskip -0.1in
    \caption{Batch Normalization statistics at various layers. TSM models are trained at 4 Frame and 16 Frame separately, and the statistics are calculated from the fourth stage of ResNet-50.}
\label{fig:more_bn}
\vskip -0.1in
\end{figure}

We have shown the calculated normalization statistics, Mean: $\mu$ and Variance: $\sigma^{2}$ in previous sections. In this part, we further include the calculated statistics of Scale: $\gamma$ and Bias: $\beta$ in Fig.~\ref{fig:more_bn}. One can observe that the two curves are not aligned with each other which further demonstrates that the discrepancy of BN statistics is an important reason for Temporal Frequency Deviation phenomenon and specializing normalization operations in deep networks is an intuitive way to resolve normalization shifting.

\subsection*{G.\quad Validation of Normalization Shifting}

\begin{figure}[h]
\vskip -0.1in
    \centering
    \begin{subfigure}[b]{0.115 \textwidth}
           \centering
           \includegraphics[width=\textwidth]{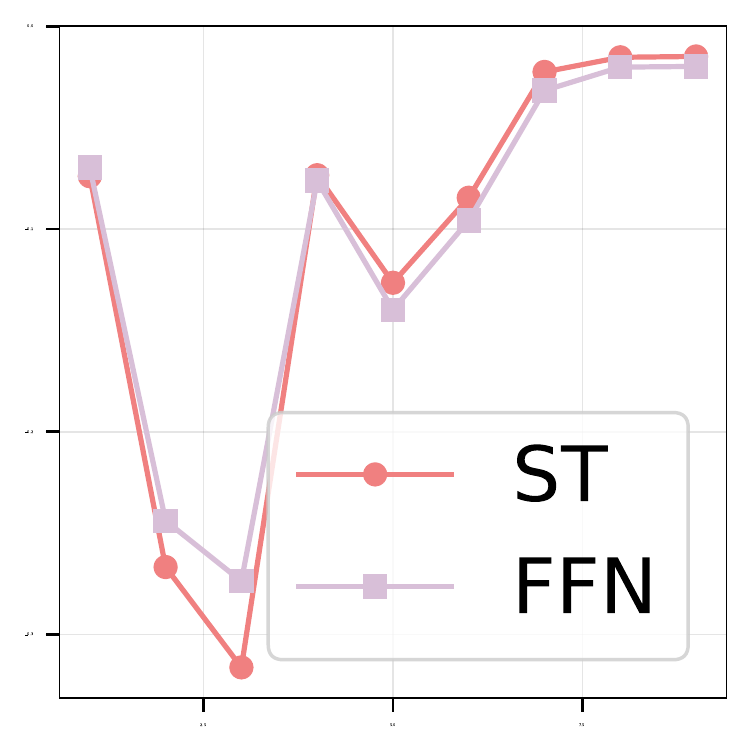}
           \vskip -0.05in
            \caption{Mean: $\mu$}
    \end{subfigure}
    \hfill
    \begin{subfigure}[b]{0.115 \textwidth}
            \centering
            \includegraphics[width=\textwidth]{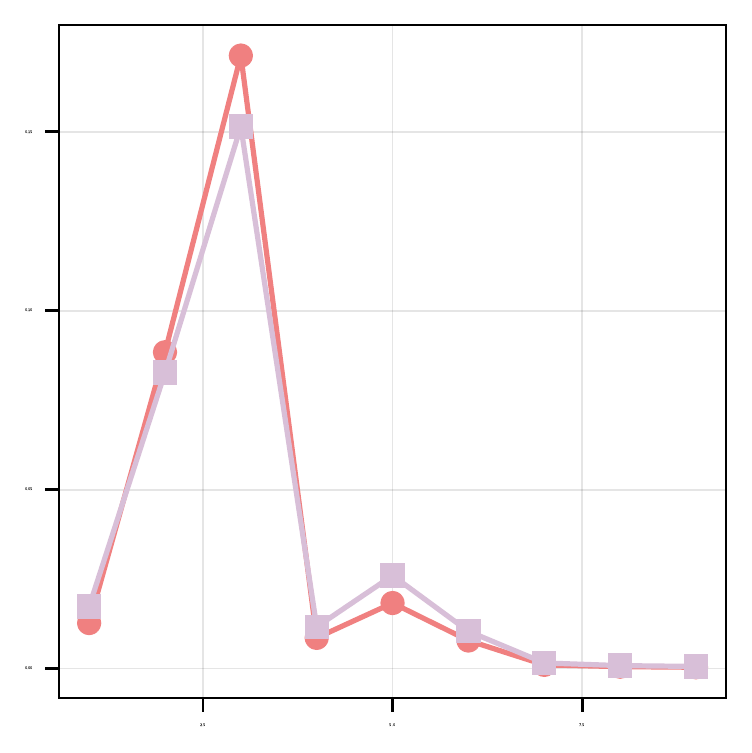}
           \vskip -0.05in
            \caption{Variance: $\sigma^{2}$}
    \end{subfigure}
    \hfill
    \begin{subfigure}[b]{0.115 \textwidth}
            \centering
            \includegraphics[width=\textwidth]{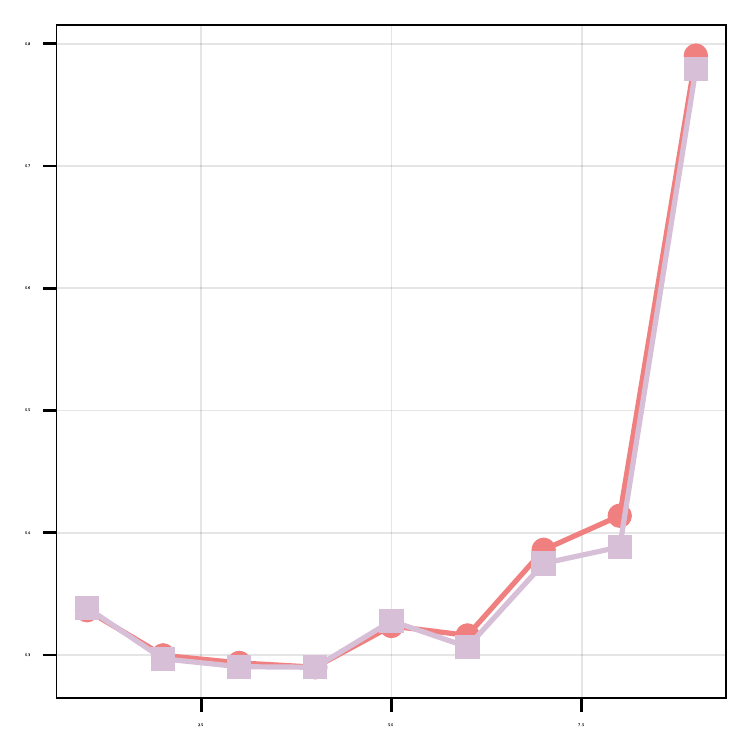}
           \vskip -0.05in
            \caption{Scale: $\gamma$}
    \end{subfigure}
    \hfill
    \begin{subfigure}[b]{0.115 \textwidth}
            \centering
            \includegraphics[width=\textwidth]{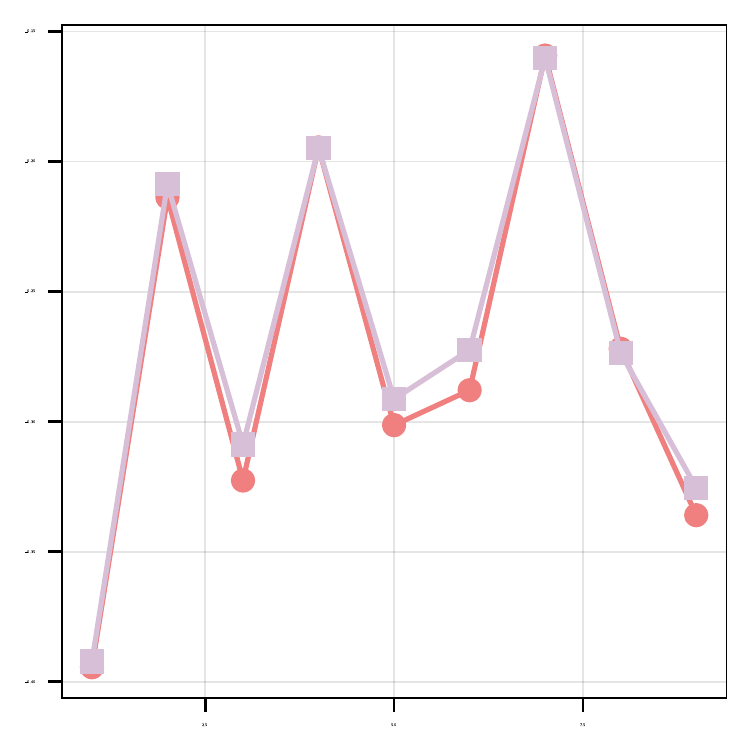}
           \vskip -0.05in
            \caption{Bias: $\beta$}
    \end{subfigure}
    \vskip -0.05in
    \caption{Batch Normalization statistics at various layers. TSM-ST is trained at 16 Frame and both models are evaluated at 16 Frame as well. The statistics are calculated from the fourth stage of ResNet-50.}
\label{fig:val_shift}
\vskip -0.1in
\end{figure}

To further prove that our method can mitigate the normalization shifting problem, we compare the BN statistics of ST (16F) and FFN (16F) which is trained with TSM~\cite{lin2019tsm} on Something-Something V1~\cite{goyal2017something} dataset. As is shown in Fig.~\ref{fig:val_shift}, one can observe that the two curves are well-aligned with each other which demonstrates that the calculated statistics are very similar and the normalization shifting problem can be alleviated by FFN.

\subsection*{H.\quad Quantitative Results}

In the Experiments section, we show performance analysis of FFN across architectures and datasets in the figure and we also provide the corresponding quantitative results in Tab.~\ref{tab:methods} and Tab.~\ref{tab:datasets} for reference.

\begin{table}[H]
\centering
\caption{Quantitative results of different architectures experiments on Something-Something V1. The best results are bold-faced.}
\vskip -0.05in
\scalebox{0.85}{\begin{tabular}{lcccc}
\toprule
\multirow{2}*{Method} & \multicolumn{3}{c}{Top-1 Acc.($\%$)} \\
\cmidrule(lr){2-5}
& $v_{L}$ & $v_{M}$ & $v_{H}$ \\
\midrule
TSM~\cite{lin2019tsm}  & 20.60 & 37.36 & 48.55 \\
TSM-ST  & 39.71 & 45.63 & 48.55 \\
TSM-FFN & \textbf{42.85}(3.14$\uparrow$) & \textbf{48.20}(2.57$\uparrow$) & \textbf{50.79}(2.24$\uparrow$) \\
\midrule
TEA~\cite{li2020tea}  & 21.78 & 41.49 & 51.23 \\
TEA-ST  & 41.36 & 48.37 & 51.23 \\
TEA-FFN & \textbf{44.97}(3.61$\uparrow$) & \textbf{51.61}(3.24$\uparrow$) & \textbf{54.04}(2.81$\uparrow$) \\
\midrule
SlowFast~\cite{feichtenhofer2019slowfast}  & 15.08 & 35.08 & 45.88 \\
SlowFast-ST  & 39.91 & 44.12 & 45.88 \\
SlowFast-FFN & \textbf{43.90}(3.99$\uparrow$) & \textbf{47.11}(2.99$\uparrow$) & \textbf{47.27}(1.39$\uparrow$) \\
\midrule
Uniformer~\cite{li2022uniformer}  & 22.38 & 47.98 & 56.71 \\
Uniformer-ST  & 44.33 & 51.49 & 56.71 \\
Uniformer-FFN & \textbf{51.41}(7.08$\uparrow$) & \textbf{56.64}(5.15$\uparrow$) & \textbf{58.88}(2.17$\uparrow$) \\
\bottomrule
\end{tabular}}
\label{tab:methods}
\end{table}

\begin{table}[H]
\centering
\caption{Quantitative results of different datasets experiments on TSM. The best results are bold-faced.}
\vskip -0.05in
\scalebox{0.75}{\begin{tabular}{lcccc}
\toprule
\multirow{2}*{Method} & \multirow{2}*{Dataset} & \multicolumn{3}{c}{Top-1 Acc.($\%$)} \\
\cmidrule(lr){3-5}
& & $v_{L}$ & $v_{M}$ & $v_{H}$ \\
\midrule
TSM~\cite{lin2019tsm}  & \multirow{3}*{Sth-Sth V2} & 31.52 & 51.55 & 61.02 \\
TSM-ST  &  & 53.38 & 59.29 & 61.02 \\
TSM-FFN &  & \textbf{56.07}(2.69$\uparrow$) & \textbf{61.86}(2.57$\uparrow$) & \textbf{63.61}(2.59$\uparrow$) \\
\midrule
TSM~\cite{lin2019tsm}  & \multirow{3}*{Kinetics400} & 64.10 & 69.77 & 73.16 \\
TSM-ST  &  & 66.25 & 70.38 & 73.16 \\
TSM-FFN &  & \textbf{68.96}(2.71$\uparrow$) & \textbf{72.33}(1.95$\uparrow$) & \textbf{74.35}(1.19$\uparrow$) \\
\midrule
TSM~\cite{lin2019tsm}  & \multirow{3}*{HMDB51} & 42.16 & 46.38 & 48.30 \\
TSM-ST  &  & 44.74 & 46.77 & 48.30 \\
TSM-FFN &  & \textbf{45.67}(0.93$\uparrow$) & \textbf{47.67}(0.90$\uparrow$) & \textbf{48.80}(0.50$\uparrow$) \\
\bottomrule
\end{tabular}}
\label{tab:datasets}
\end{table}

\end{document}